\documentclass[10pt,twocolumn,letterpaper]{article}
\usepackage{cvpr}
\usepackage{times}

\usepackage[utf8]{inputenc}
\usepackage{amsmath}
\usepackage{amssymb}
\usepackage{graphicx}
\usepackage{epsfig}
\usepackage{subcaption}
\usepackage{romannum}
\usepackage{multicol}
\usepackage{multirow}
\usepackage{array}
\usepackage{tabu}
\usepackage{url}

\usepackage[noblocks]{authblk}

\usepackage[tableposition=top]{caption}
\usepackage{float}
\newcommand{\HI} {\color{black}}
\newcommand{\bck} {\color{black}}
\newcommand{\js} {\color{black}}

\usepackage[breaklinks=true,bookmarks=false]{hyperref}

\cvprfinalcopy % *** Uncomment this line for the final submission

\def\cvprPaperID{****} % *** Enter the CVPR Paper ID here

\begin{document}

%%%%%%%%% TITLE
\title{Multi-Scale Gradual Integration CNN \\ for False Positive Reduction in Pulmonary Nodule Detection}

\author[1]{Bum-Chae~Kim}
\author[1]{Jun-Sik Choi}
\author[1]{Heung-Il Suk}
\affil[1]{Department of Brain and Cognitive Engineering, Korea University, Seoul, Republic of Korea}

\maketitle

\begin{abstract}
Lung cancer is a global and dangerous disease, and {\HI its} early detection is crucial to reducing the risks of mortality. {\HI In this regard, it has been} of great interest in developing a {computer-aided system} for {\HI pulmonary nodules detection as early as possible} on thoracic CT scans. {\HI In general, a} nodule detection system involves two steps: (i) candidate nodule detection at a high sensitivity, which captures many false positives and (ii) false positive reduction from candidates. However, due to the high variation of nodule morphological characteristics and the possibility of mistaking them for neighboring organs, candidate nodule detection remains a challenge. In this study, we propose a novel {Multi-scale Gradual Integration Convolutional Neural Network (MGI-CNN)}, designed with three main strategies: (1) to use multi-scale inputs with different levels of contextual information, (2) to use abstract information inherent in different input scales with gradual integration, and (3) to learn multi-stream feature integration in an end-to-end manner. To verify the efficacy of the proposed network, we conducted exhaustive experiments on the LUNA16 challenge datasets by comparing the performance of the proposed method with state-of-the-art methods in the literature. On two candidate subsets of the LUNA16 dataset, \ie, V1 and V2, our method achieved an average CPM of 0.908 (V1) and 0.942 (V2), outperforming comparable methods by a large margin. Our MGI-CNN is implemented in Python using TensorFlow and the source code is available from \url{https://github.com/ku-milab/MGICNN}.
\end{abstract}

%%%%%%%%% BODY TEXT
\section{Introduction}
% importance of automatic pulmonary nodule detection
Lung cancer is reported as the leading cause of death worldwide \cite{Siegel2017Cancer2017}.
However, when detected at an early stage through thoracic screening with low-dose CT images and treated properly, the survival rate can be increased by 20\% \cite{NationalLungScreeningTrialResearchTeam2011ReducedScreening}. 
{\HI Clinically, pulmonary nodules are characterized as having round shape with a diameter of $3mm\sim30mm$ in thoracic CT scans \cite{Gould2007EvaluationEdition}. With this pathological knowledge, there have been efforts of applying machine-learning techniques for early and automatic detection of cancerous lesions, \ie, nodules.} {\HI To our knowledge}, a computerized lung cancer screening system consists of two-steps: candidate nodule detection and False Positives (FPs) reduction. 
In the candidate nodule detection step, the system uses high sensitivity without concern for specificity to extract as many candidates as possible. {\HI Roughly,} \textcolor{black}{more than 99\%} of the candidates are non-nodules, \ie, FPs \cite{Setio2016PulmonaryNetworks}, which {\HI should} be identified and reduced in the second step {\HI correctly}.

\begin{figure*}[t] \centering
\begin{subfigure}{0.49\textwidth} \centering
\includegraphics[width=0.22\linewidth, height=1.3cm]{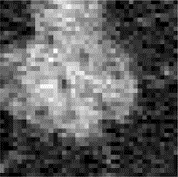}
\includegraphics[width=0.22\linewidth, height=1.3cm]{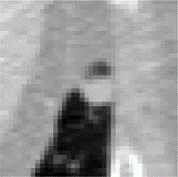}
\includegraphics[width=0.22\linewidth, height=1.3cm]{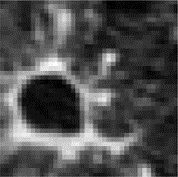}
\includegraphics[width=0.22\linewidth, height=1.3cm]{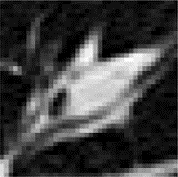}
\includegraphics[width=0.22\linewidth, height=1.3cm]{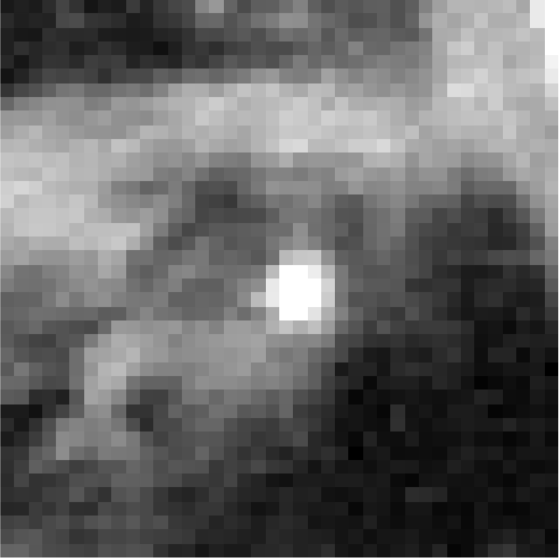}
\includegraphics[width=0.22\linewidth, height=1.3cm]{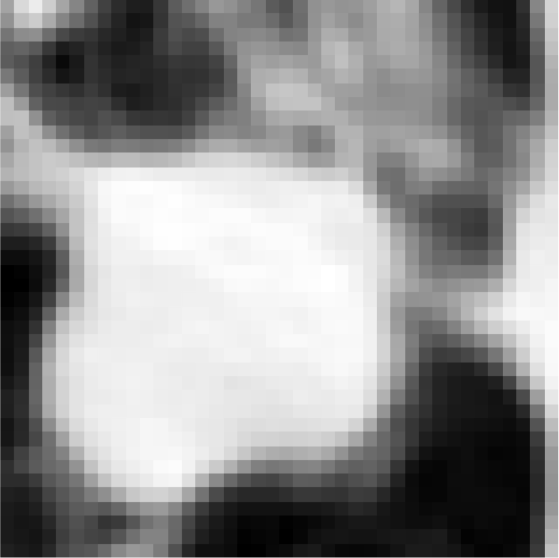}
\includegraphics[width=0.22\linewidth, height=1.3cm]{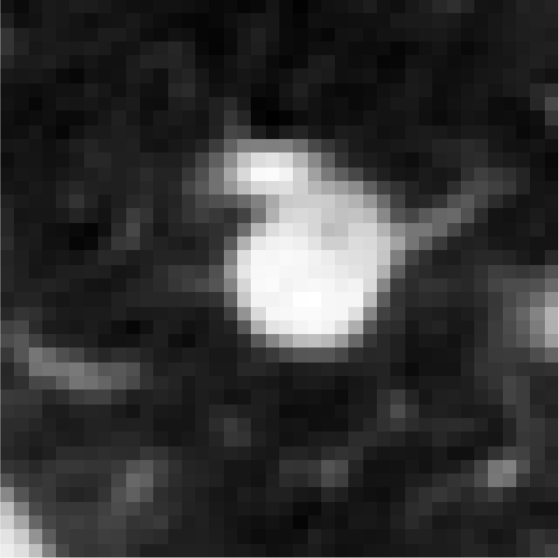}
\includegraphics[width=0.22\linewidth, height=1.3cm]{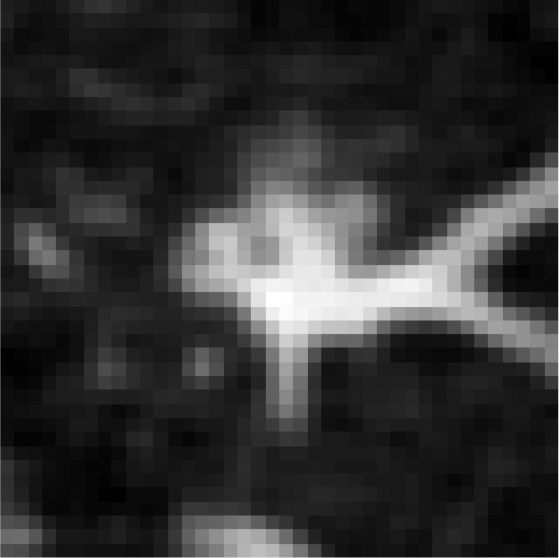}
\caption{Nodule}\label{fig:nd_size}
\end{subfigure}
\begin{subfigure}{0.49\textwidth} \centering
\includegraphics[width=0.22\linewidth, height=1.3cm]{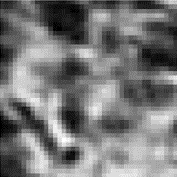}
\includegraphics[width=0.22\linewidth, height=1.3cm]{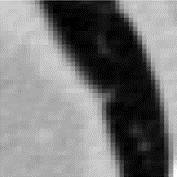}
\includegraphics[width=0.22\linewidth, height=1.3cm]{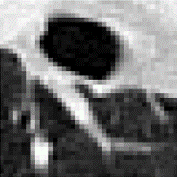}
\includegraphics[width=0.22\linewidth, height=1.3cm]{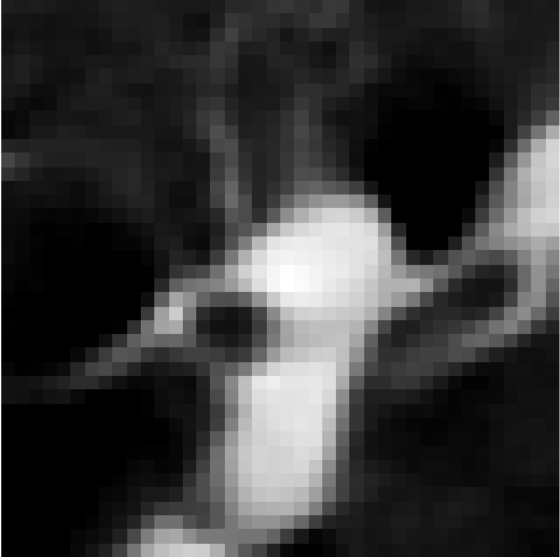}
\includegraphics[width=0.22\linewidth, height=1.3cm]{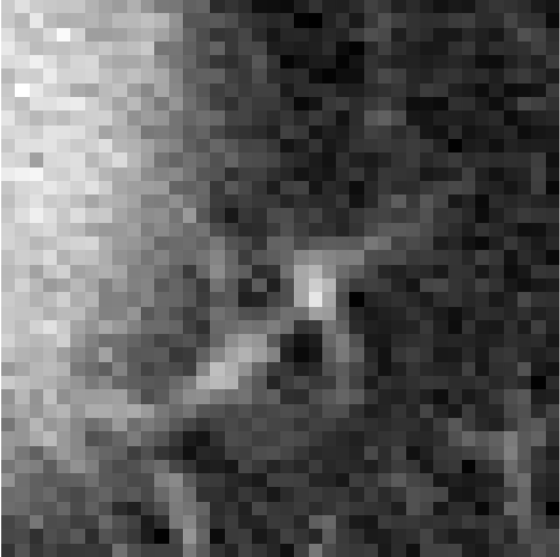}
\includegraphics[width=0.22\linewidth, height=1.3cm]{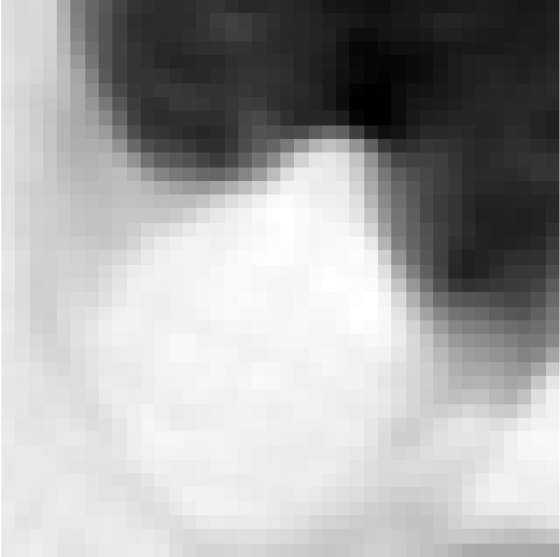}
\includegraphics[width=0.22\linewidth, height=1.3cm]{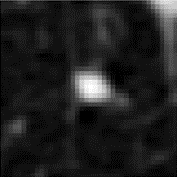}
\includegraphics[width=0.22\linewidth, height=1.3cm]{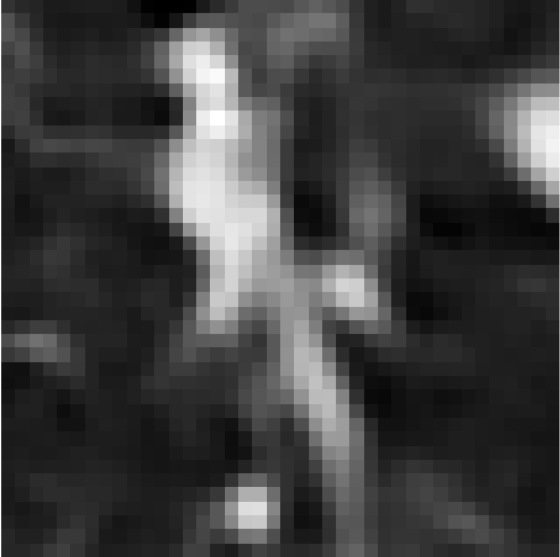}
\caption{Non-nodule}\label{fig:nnd_size}
\end{subfigure}
\caption{Examples of the pulmonary nodules and non-nodules. Both have a complex and similar morphological characteristics {\HI that must be distinguished between}. 
\label{fig:image1}}
\end{figure*}

{\HI Pathologically,} there are many types of nodules {\bck (\eg, solids, non-solids, part-solids, calcified, \etc. \cite{Ciompi2017TowardsLearning})} and {\HI their} morphological characteristics such as size, shape, and strength are {\HI highly} variable. In addition, there are many other structure in the thorax (\eg, blood vessels, airways, lymph nodes) with morphological features similar to nodules \cite{Gould2007EvaluationEdition, Roth2016ImprovingAggregation}. {\HI Fig.} \ref{fig:image1} shows an example of a nodule and a non-nodule. {\HI In these regards, it is very challenging to reduce FPs or to distinguish nodules from non-nodules, leading many researchers to devote their efforts on the step of false positive reduction} \cite{Setio2016PulmonaryNetworks, Dou2016Multi-levelDetection, CAO2017327}.

In earlier work, researchers had {\HI mostly focused on} extracting discriminative morphological features with the help of {\HI pathological} knowledge about nodule types {\HI and applied relatively simple linear classifiers such as logistic regression or support vector machine} \cite{Lee2001AutomatedTechnique, Okumura1998AutomaticFilter, Ye2009Shape-basedImages}. {\HI Recently, with the surge of popularity and success in Deep Neural Networks (DNNs)}, which can learn hierarchical feature representations and class discrimination in a single framework, a myriad of DNNs has been proposed for medical image analysis \cite{Suzuki2003MassiveTomography, Dou2016AutomaticNetworks, Esteva2017Dermatologist-levelNetworks, Havaei2017BrainNetworks, Shen2017DeepAnalysis., HU2018134}.
% In the mean time, according to the recent surge of popularity and successes of deep neural networks, which can learn hierarchical feature representations and a class discriminant in a single framework, for computer vision tasks, a myriad of deep models have also been proposed in medical image analysis \cite{Esteva2017Dermatologist-levelNetworks, Havaei2017BrainNetworks, Shen2017DeepAnalysis.}.
Of the various {\HI deep models}, Convolutional Neural Networks (CNNs) have been applied most successfully for pulmonary nodule detection and classification in CT images \cite{Murphy2009AClassification, Jacobs2014AutomaticImages, Setio2015AutomaticImages, Setio2016PulmonaryNetworks, Roth2016ImprovingAggregation, LIU2018262, SHEN2017663}. 
% Of various deep neural networks, Convolutional Neural Networks (CNNs) have been the most successfully applied for pulmonary nodule detection or classification in CT images \cite{Setio2016PulmonaryNetworks, Roth2016ImprovingAggregation, Jacobs2014AutomaticImages, Murphy2009AClassification, Setio2015AutomaticImages}. 
{\HI Moreover, in order to} attain the network performance of computer vision applications, there were trials \cite{Ciompi2015AutomaticOut-of-the-box, Shin2016DeepLearning} {\HI to identify nodules with a deep model fine-tuned with} pulmonary nodule data  {\HI in the way of transfer learning \cite{Yang2009_TL, Razavian2014}}.
% To utilize the network performance used in the computer vision, the classify nodules by fine tuning model with pulmonary nodule data \cite{Ciompi2015AutomaticOut-of-the-box, Shin2016DeepLearning}. 
% However, the performances were not better than hand-crafted filters. 

%Recently, automatic pulmonary detection using deep learning models has emerged. Among them, CNN shows good performance in medical image analysis  
From previous studies of nodule detection or classification in CT scans, we have two notable findings. 
The first is that it is helpful to exploit volume-level information, {\HI rather than 2D slice}-level information \cite{Roth2016ImprovingAggregation,Setio2016PulmonaryNetworks,Ding2017AccurateNetworks}.
For example, Roth \etal~\cite{Roth2016ImprovingAggregation} proposed a 2.5D CNN by taking three orthogonal 2D patches as input for volume-level feature representation. Setio \etal~\cite{Setio2016PulmonaryNetworks} proposed a multi-view CNN, which extracts hierarchical features from nine 2D slices with different angles of view, and groups the high-level features for classification. However, their method achieved limited performance in low-FP scans. Ding \etal~\cite{Ding2017AccurateNetworks} proposed a 3D CNN with a 3D volumetric patch as input, and presented promising results in FP reduction.
% From the previous studies of nodule detection or classification in CT, we learn two notable findings. 
% First, it is helpful to exploit volume-level information, instead of 2D image-level information \cite{Roth2016ImprovingAggregation,Setio2016PulmonaryNetworks,Ding2017AccurateNetworks}. 
% For example, Roth \etal \cite{Roth2016ImprovingAggregation} proposed a 2.5D CNN by taking three orthogonal 2D patches as input for volume-level feature representation. 
% Setio \etal \cite{Setio2016PulmonaryNetworks} proposed a multi-view CNN, which extracts hierarchical features from nine 2D slices with different view-angles and ensembles the high-level features for classification. However, their method showed limited performance in low FP/scan. Ding \etal proposed a 3D-CNN with a 3D volumetric patch as input and presented promising results in false positive reduction.

The second is that performance can be enhanced by using multi-scale inputs with different levels of contextual information \cite{Shen2015,Dou2016Multi-levelDetection,Shen2017Multi-cropClassification}. Shen \etal~\cite{Shen2015} proposed a multi-scale CNN and successfully applied nodule classification by combining contextual information at different image scales with the {\HI abstract-level feature representations}. 
Dou \etal~\cite{Dou2016Multi-levelDetection} also designed a 3D CNN to encode multi-level contextual information to tackle the challenges of large variation in pulmonary nodules. 
The performance of pulmonary nodule classification using the 3D CNN is {\HI generally} better than that of the 2D CNN \cite{Ding2017AccurateNetworks}. However, the 3D CNN is more difficult to train than the 2D CNN due to the large number of network parameters. Medical image data is relatively limited, so a 3D CNN may easily become over-fitted. It is also noteworthy that the multi-scale methods have proved their efficacy in computer vision tasks \cite{Karpathy2014Large-ScaleNetworks, Honari2016RecombinatorAggregation, Lin2016FeatureDetection}.

Inspired by the above-mentioned findings, in this study we propose a novel \textcolor{black}{Multi-scale Gradual Integration CNN (MGI-CNN)} for FP reduction in pulmonary nodule detection. In designing our network, we apply three main strategies. Strategy 1: We use 3D multi-scale inputs, each containing different levels of contextual information. Strategy 2: We design a network for Gradual Feature Extraction (GFE) from multi-scale inputs at different layers, instead of radical integration at the same layer  \cite{Karpathy2014Large-ScaleNetworks, Shen2015, Shen2017Multi-cropClassification, Dou2016Multi-levelDetection}. Strategy 3: For better use of complementary information, we consider \textcolor{black}{Multi-Stream Feature Integration (MSFI) to integrate abstract-level feature} {\HI representations}. 
Our {\HI main} contributions can be summarized as follows:

\begin{enumerate}
\item We propose a novel CNN architecture that learns feature representations of multi-scale inputs with a gradual feature extraction strategy.
% We propose a novel CNN architecture that learns feature representations of multi-scale inputs with a gradual feature integration strategy.
% complementary context information from 3D spatial region by using two streams subnetwork in GFE method.
%\item With MSFIs and abstract feature integration, our network can greatly reduce FPs.
\item {\HI With multi-stream feature representations and abstract-level feature integration, our network reduces many false positives.}
\item Our method outperformed state-of-the-art methods in the literature by a large margin on the LUNA16 challenge {\HI datasets}.
% Our method outperformed the state-of-the-art methods in the literature with large margin on the LUNA16 challenge dataset.% other published methods on the average Competition Performance Metric (CPM) which is used in the LUNA16 challenge, and in particular, our method achieved high performance especially when the FP per scan is low.
\end{enumerate}

\textcolor{black}{While the proposed network architecture extension is straightforward, to our best knowledge, this is the first work of designing a network architecture that integrates 3D contextual information of multi-scale patches in a gradual and multi-stream manner. Concretely, our work empirically proved the validity of integrating multi-scale contextual information in a gradual manner, which can be comparable to many existing work  \textcolor{black}{\cite{Lin2016FeatureDetection,Kamnitsas2017EfficientSegmentation}} that mostly considered radical integration of such information. Besides, our method also presents the effectiveness of learning feature representations from different orders of multi-scale 3D patches and combining the extracted features from different streams to further enhance the performance.}

% {\bc We used an intuitive network architecture extension method to extract and integrate features extracted from a multi-scale patch into different streams. 
% However, to our best knowledge, this is the first time we have proposed a network architecture designed to integrate 3D context information from a multi-scale patch into a gradually multi-stream. Concretely, our study empirically proved the feasibility of integrating multi-dimensional contextual information in a gradual way, which can be compared with many previous studies \cite{Lin2016FeatureDetection,Kamnitsas2017EfficientSegmentation}, which are primarily concerned with the radical integration of such information. In addition, our method also has the effect of further enhancing performance by learning feature expressions from multi-scale 3D patches of different dimensions and combining extracted features of different streams.}

This paper is organized as follows. Section \ref{sec:related_work} introduces the existing methods in the literature. We then describe our proposed method in Section \ref{sec:proposed_method}. The experimental settings and performance comparison with the state-of-the-art methods are presented in Section \ref{sec:experiemtns}. In Section \ref{sec:discussion}, we discuss key issues of the proposed method along with the experimental results. We conclude this paper by summarizing our work and suggesting the future direction for clinical practice in Section \ref{sec:conclusion}.

% The paper is organized as follows. Section \ref{sec:related_work} introduces the existing methods in the literature. We then describe our proposed method in Section \ref{sec:proposed_method}. The experimental settings and performance comparison with the state-of-the-art methods are presented in Section \ref{sec:experiemtns}. In Section \ref{sec:discussion}, we discuss some key issues of the proposed method along with the experimental results. We conclude this paper by summarizing of our work and suggesting the future direction for clinical practice in Section \ref{sec:conclusion}.

\section{Related Work}
\label{sec:related_work}

\subsection{Volumetric Contextual Information}
Automatic lung cancer screening systems classify nodules using specific algorithms to extract {\HI nodule} morphological characteristics. 
Okumura \etal,~\cite{Okumura1998AutomaticFilter} {\HI distinguished} solid nodules by using a Quoit filter that could detect only isolated nodules. In the case of isolated nodules, the graph of the pixel values becomes `sharp,’ and the nodule is detected when the annular filter passes through the graph. However, filters that use only one characteristic of nodules have difficulty {\HI in} distinguishing diverse nodule types. 
% Okumura \etal~\cite{Okumura1998AutomaticFilter} wanted to distinguish solid-nodule by using a `Quoit' filter that can detect only the isolated nodules. In the case of isolated nodules, the graph of the pixel values becomes `sharp', and the nodule can be detected when the annular filter passes through the graph. However, the filters that only use the one characteristic of nodules are difficult to distinguish the diverse type nodules.
Li \etal~\cite{Li2003SelectiveScans} proposed point, line, and surface shape filters for finding nodule, blood vessel, and airway in a thoracic CT. This is a detection method that considers various types of nodules, effectively reducing the FP response of the automatic lung cancer screening system. However, hand-crafted features still do not detect complex types of nodules (\eg, part-solid {\HI or calcified} nodules). Hence, to detect the more elusive types of nodule, researchers attempted to use volumetric information about the nodule and {\js its} surrounding area. 
% Li \etal~\cite{Li2003SelectiveScans} proposed point, line, and surface shape filters for finding nodule, blood vessel, and airway in a thoracic CT. This is a method for detecting various types of nodules taking into consideration, effectively reducing the false positive response of the automatic pulmonary nodule detection system.
% However, the hand-crafted features still did not detect the complex type of nodules (\eg, part-solid nodule). Hence, to detect the difficult type of nodule, researchers attempted the volumetric information of the nodule and surrounding area.
Jacobs \etal~\cite{Jacobs2014AutomaticImages} extracted volumetric information from various types of bounding boxes that defined the region around a nodule to classify \textcolor{black}{part-solid} nodules. That volumetric information includes 107 phenotype features and 21 context features of the nodule and various nodule area with {\HI diverse sizes of a} bounding box. For the classification, the GentleBoost classifier \cite{Friedman1998AdditiveBoosting} learned a total of 128 features {\js and obtained 80\% of sensitivity at 1.0 FP/scan.} However, the method was inefficient in distinguishing the various types of nodule because it must be reconfigured to filter each nodule type.
Recently, {\HI DNNs have been successfully used to substitute the conventional pattern-recognition approaches that first extract features and then train a classifier separately, thanks to their ability of discovering data-driven feature representations and training a classifier in a unified framework. Among various DNNs, CNN-based methods reported promising performance in classifying nodules correctly.}
%\textcolor{red}{CNNs} has shown particularly good performance in the image screening specialist assistant system. 
%Because a CNN algorithm is based on the human visual system, CNN observes the input image and automatically extracts all the volumetric information. CNNs do not simply observe nodules from a single point of view, but rather, they observe nodules from various locations to extract high-level features necessary for classification. 
%The performance of the classification method improves through the grouping of these high-level features.
% Recently, Deep Neural Network (DNN) show the state-of-the-art performance in computer vision and natural language process. In the field of medical image analysis, DNN method was used as a substitute for image screening specialist. In particular, CNN is algorithmized based on the human visual system, CNN observes the input image and automatic extracts all the volumetric information. 
% This information may be further classified because it includes not only human-viewable features but also network-observed features. 
Roth \etal~\cite{Roth2016ImprovingAggregation} proposed 2.5D CNN that used three anatomical planes (sagittal, coronal, and axial) to extract 3D nodule area volumetric information. Their 2.5D CNN also classified organs similar to nodules, such as lymph nodes. This study inspired some researchers in the field of pulmonary nodule detection.
% Roth \etal~\cite{Roth2016ImprovingAggregation} proposed 2.5D CNN which uses three anatomical planes (sagittal, coronal and axial) to extract 3D nodule area volumetric information. This study inspired some researchers in pulmonary nodule automatic detection.
Setio \etal~\cite{Setio2016PulmonaryNetworks} proposed a multi-view CNN that extracted volumetric information with an increased number of input patches. Furthermore, to better consider contextual information, they used groupings of high-level features from each 2D CNN in a 9-view (three times more than 2.5D CNN's anatomical plane) {\HI by achieving promising performance, compared with the methods using hand-crafted features}.
However, this effort could not {\HI fully utilize} all the 3D volumetric information {\HI that could be useful to further enhance the performance.} 
% Setio \etal~\cite{Setio2016PulmonaryNetworks} proposed a multi-view CNN that extracts volumetric information with more input patches. Furthermore, to consider of contextual information in 9-views, they use ensembles high-level features from each 2D CNN in a 9-views. However, this effort to extract 3D volumetric information could not the extract all of the information because of the limitations of the 2D patch.
Ding \etal~\cite{Ding2017AccurateNetworks} {\HI tried to build a unified framework by applying a deep CNN for both candidate nodule detection and nodule identification. Specifically, they designed a deconvolutional CNN structure for candidate detection on axial slices and a three-dimensional deep CNN for the subsequent FP reduction. In the FP reduction step, they used a dropout method by achieving a sensitivity of 0.913 in average FP/scan on the LUNA16 dataset. Although they claimed to use 3D volumetric information, they did not consider the information between the small patches that were extracted in a large patch.}
%To improve the classification performance of nodules, they used a dropout layer with the end of convolution layers. To demonstrate the efficiency of a network by achieving an average 0.913 FP/scan sensitivity on the LUNA16 dataset. Although they claimed to use whole 3D volumetric information, they did not consider the information between the small patches that were extracted in a large patch.

% Ding \etal~\cite{Ding2017AccurateNetworks} use 3D patches as input to volumetric information. They use 3D sub-patches which decompose a large 3D patch allow overlapping. To improve the classification performance of nodules, they used a dropout to demonstrate efficiency by achieving 0.913 sensitivity in average FP/scan LUNA16 dataset. However, They claim to use 3D volumetric information, they did not consider the information between the small patches that were extracted in a large patch. 

\subsection{Multi-scale Contextual Information}
%The 3D CNN is the most suitable method to extract information from 3D nodule images, but the number of learning data can not keep up with the increase in the number of parameters in the network. \textcolor{red}{Therefore, researcher use multi-scale patches to extract diverse volumetric information of nodules.}
{\HI From an information quantity perspective, it may be reasonable to use morphological and structural features in different scales and thus effectively integrating multi-scale contextual information.} 
Shen \etal~\cite{Shen2015} proposed Multi-scale CNN (MCNN) as a method for extraction of high-level features from a single network by converting images of various scales to the same size. The high-level features are {\HI jointly} used to {\HI train a classifier{\js ,} such as support vector machine or random forest,} for nodule classification.
% Shen \etal~\cite{Shen2015} proposed Multi-scale CNN (MCNN) is a method of learning Support Vector Machine (SVM) and Random Forest (RF) classifier by connecting high-level features extracted from a single network by changing images of various scales to the same size.
Dou \etal~\cite{Dou2016Multi-levelDetection} used three {\HI different} architectures of 3D CNN\HI, each one of which was trained with the respective receptive field of an input patch empirically optimized for the LUNA16 challenge dataset. To make a final decision, they integrated label prediction values from patches of three different scales by a weighted sum at the top layers. However, the weights for each scale were determined manually, rather than learning from training samples.
%The size of the receptive fields is determined by the ratio of the number of nodules. To fully train of 3D CNN, they used three networks consisting of seven layers in total. One of these seven layers is the Softmax layer and consists of two fully connected layers. There are three convolution layers for extracting features from the image, and Max Pooling also has one layer. The features of the nodule were extracted through this network, and weighted predictions were used for the final classification. 
%The network is too small to extract sufficient features to classify nodules and nodules. In addition, even if good features are extracted through sufficient learning, it is difficult to utilize the network ensembles by weighting them using the values specified by the researcher.

% Dou \etal~\cite{Dou2016Multi-levelDetection} used three 3D CNNs with three scale of 3D patches. the patch sizes are determined by the ratio including the nodules. To determine the final predicted value of candidate nodules weights are arbitrarily multiplied by the predicted value of the network.
%Dou \etal~\cite{Dou2016Multi-levelDetection} used three architecture of 3D CNN with optimized receptive fields. The size of the receptive fields are determined by the ratio including the nodules, and the weights of each receptive fields are arbitrarily multiplied by the predicted value of the network to determine the final predicted value of candidate nodules. 
{\HI Multi-crop Convolutional Neural Network (MC-CNN) to automatically extract nodule salient information by employing a novel multi-crop pooling strategy which crops different regions from convolutional feature maps and then applies max-pooling different times.}

Shen \etal~\cite{Shen2017Multi-cropClassification} proposed a Multi-Crop CNN {\HI to automatically extract nodule salient information by employing a novel multi-crop pooling strategy. In particular, they cropped different regions from convolutional feature maps and then applied a max-pooling operation different times. To give more attention on the center of the patches, they cropped out the neighboring or surrounding information during multi-crop pooling, which could be more informative to differentiate nodules from non-nodules, \eg, other organs.}

{\HI 
In this paper, unlike the methods of \cite{Setio2016PulmonaryNetworks, Roth2016ImprovingAggregation}, we exploit 3D patches to best utilize the volumetric information and thus enhancing the performance in FP reduction. Further, to utilize contextual information from different scales, we exploit a multi-scale approach similar to \cite{Dou2016Multi-levelDetection, Shen2017Multi-cropClassification}. However, instead of radical integration of multi-scale contextual information at a certain layer \cite{Shen2015,Shen2017Multi-cropClassification}, we propose to gradually integrate such information in a hierarchical fashion. It is also noteworthy that we still consider the surrounding regions of a candidate nodule to differentiate from other organs, which can be comparable to \cite{Shen2017Multi-cropClassification}.
}

\begin{figure*}[t] \centering
\begin{subfigure}{0.45\textwidth}\centering
\includegraphics[width=\textwidth]{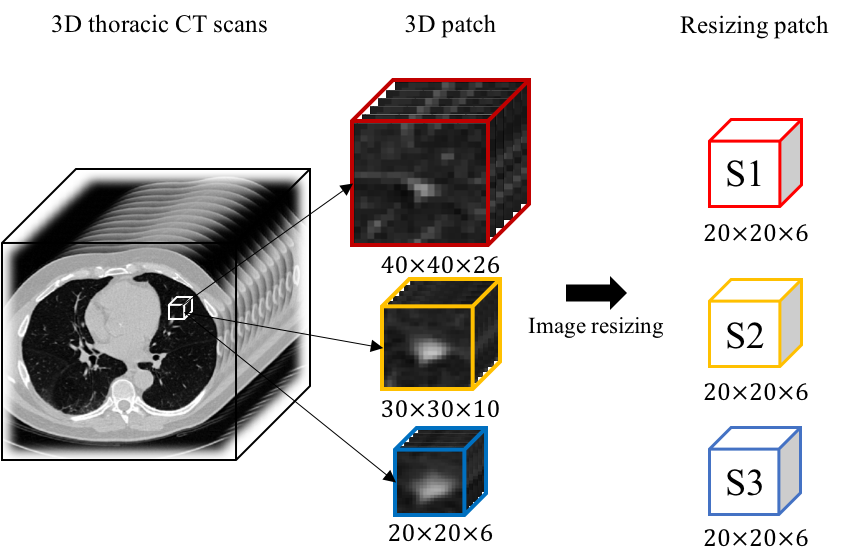}
\caption{3D Patch Extraction: Given the coordinates of a candidate nodule, we extract three patches in different scales and then resize them to the same size, \ie, $S1$, $S2$, and $S3$.\label{fig:3dpatchextraction}}
\end{subfigure}
\begin{subfigure}{0.45\textwidth}\centering
\includegraphics[width=\textwidth]{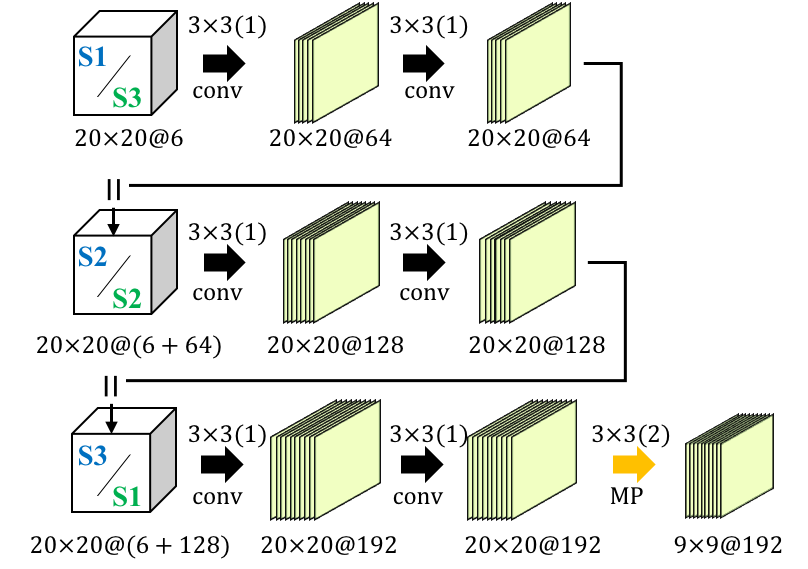}
\caption{Gradual Feature Extraction: The multi-scale patches with different levels of contextual information ($S1$-$S2$-$S3$ or $S3$-$S2$-$S1$) are integrated in a gradual manner. 
%{\bck The abstracted information through Max pooling (MP) is finally extracted.} {\color{green}no pooling?}
\label{fig:GFE}}
\end{subfigure}
\begin{subfigure}{\textwidth}\centering
\includegraphics[width=\textwidth]{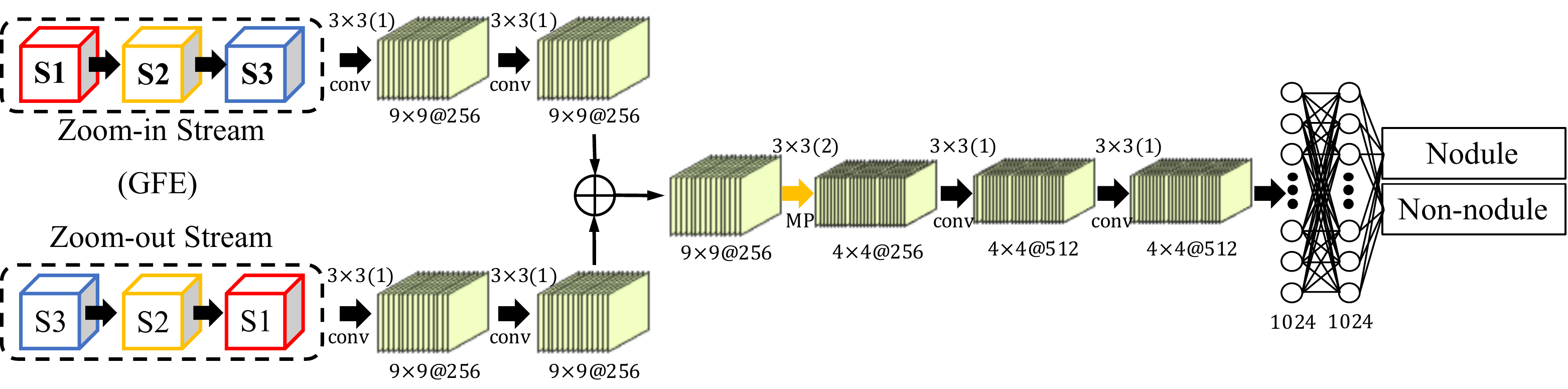}
\caption{The architecture of the proposed multi-scale gradual integration CNN. %{\bck Our proposed network use  $3\times3$ convolution kernel with 1 stride and zero-padding. Max pooling, for information of abstract, is used twice, \ie, the end layer of GFE and the behind of element-wise summation.}{\color{green}no pooling?} 
\label{fig:MSFR}}
\end{subfigure}
\caption{Overview of the propose framework for FP reduction in pulmonary nodule detection. The notations of $\Vert$ and $\oplus$ denote, respectively, concatenation and element-wise summation of feature maps. The numbers above the thick black or yellow arrows present a kernel size, \eg, $3\times3$ and a stride, \eg, (1) and (2). (conv: convolution, MP: max-pooling)
} %(a) Given the coordinate of a nodule candidate, we extract three patches in different scales and then resize them to the same size, \ie, $S1$, $S2$, and $S3$. (b) The multi-scale patches ($S1$-$S2$-$S3$ or $S3$-$S2$-$S1$) are then integrated in a hierarchical manner via CNN. (c) The final architecture of the proposed multi-scale gradual integration CNN for FP reduction in pulmonary nodule detection.}
\label{fig:proposed}
\end{figure*}

\section{Multi-scale Gradual Integration Convolutional Neural Network (MGI-CNN)}
\label{sec:proposed_method}
{\HI In this section, we describe our novel method of Multi-scale Gradual Integration Convolutional Neural Network (MGI-CNN) in Fig. \ref{fig:proposed} for pulmonary nodule identification, which consists of two main components: Gradual Feature Extraction (GFE) and Multi-Stream Feature Integration (MSFI).} 
%networks for FP reduction in lung cancer screening system is shown in Fig. \ref{fig:proposed}. 
\textcolor{black}{For each candidate nodule, we extract 3D patches at three different scales  $40\times40\times26$, $30\times30\times10$, and $20\times20\times6$ {\HI by following} Dou \etal's work \cite{Dou2016Multi-levelDetection}. We then resize three patches to $20\times20\times6$, {\HI, denoted as $S1$, $S2$, and $S3$, respectively, as input to the proposed network (Fig. \ref{fig:3dpatchextraction}). Note that patches of $S1$, $S2$, and $S3$ have the same center coordinates but pixels in the patches are different in resolution.}}

\subsection{Gradual Feature Extraction}
Inspired by the human visual system, which retrieves meaningful contextual information from a scene by changing the field of view, \ie, {\HI by considering contextual information at multiple scales} \cite{Zhang2014ScaleAnalysis}, we first propose a scale-ordered GFE network {\HI presented in Fig. \ref{fig:GFE}. 
%{\HI In this paper, we consider three different scales and let us denote the patches $S1$, $S2$, and $S3$ from smallest to largest in scales.}
In integrating morphological or structural features from patches at different scales, the existing methods \cite{Shen2015,Shen2017Multi-cropClassification} combined features from multiple patches all at once. Unlike their methods, in this paper, we extract features by gradually integrating contextual information from different scales in a hierarchical manner.} 
For gradual feature representation from multi-scale patches, \ie, $S1$, $S2$, and $S3$, there are two possible scenarios, \ie, $S1-S2-S3$ (`\emph{zoom-in}') or $S3-S2-S1$ (`\emph{zoom-out}'). 
% There are two scenarios for using the features of the multi-scale patches, \ie, $S1-S2-S3$ (`\emph{zoom-in}') or $S3-S2-S1$ (`\emph{zoom-out}').
% Inspired by the human visual system that retrieves meaningful information from a scene by changing the field of view, \ie, multiple scales %, where it alternates moving closer to the object of interest and moving further away from it 
% \cite{Zhang2014ScaleAnalysis}, we propose a scale-ordered \emph{Gradual Feature Extraction (GFE)} network (Fig. \ref{fig:GFE}). 
% For feature representation from multi-scale patches, \ie, $S1$, $S2$, and $S3$, there are two possible scenarios, \ie, $S1-S2-S3$ (`\emph{zoom-in}') or $S3-S2-S1$ (`\emph{zoom-out}'). 

For the zoom-in scenario of $S1-S2-S3$, a patch at one scale $S1$ is first filtered by the corresponding local convolutional kernels and the resulting feature maps $F1$ are concatenated ($\Vert$) with the patch at the next scale $S2$, \ie, $F1 \Vert S2$. In our convolution layer, $F1$ is the result of {\HI two repeated computations of a spatial convolution and a non-linear transformation by a Rectifier Linear Unit (ReLU) \cite{Nair2010RectifiedMachines}}. Our convolution kernel uses zero padding to keep the size of the output {\HI feature maps equal to the size of an input patch and thus valid to concatenate the resulting feature maps and another input patch $S2$ in different scale}.
The $F1 \Vert S2$ tensor is then convolved with kernels of the {\HI following} convolution layers, producing feature maps $F12$, which now represent the integrated contextual information from $S1$ and $S2$. The feature maps $F12$ are then concatenated with the patch at the next scale $S3$ and the tensor of $F12 \Vert S3$ is processed by the related kernels, resulting in feature maps $F123$. The feature maps $F123$ represent the final integration of various types of contextual information in patches $S1$, $S2$, and $S3$. {\HI The number of feature maps in our network increases as additional inputs} are connected so that the additional information can be extracted from the information of the preceding inputs and the contextual information of the sequential inputs.
For the zoom-out scenario of $S3-S2-S1$, the same operations are performed but with input patches in the opposite order.
%our network to combine three scale-ordered use a 3D CNN for complex morphological features from nodules and non-nodules. 

In the zoom-in scenario, the network is provided with patches at an increasing scale. So, the field of view in a zoom-in network is gradually reduced, meaning that the network gradually focuses on a nodule region. Meanwhile, the zoom-out network has a gradually enlarging field of view, and thus the network finds morphological features combined with the neighboring contextual information by gradually focusing on the surrounding region. In our network architecture, the feature maps extracted from the previous scale are concatenated to the patch of the next scale with zero padding, and then fed into the following convolution layer.
By means of our GFE method, our network sequentially integrates contextual features according to the order of the scales. 
{\HI It is noteworthy that the abstract feature representations from two different scenarios, \ie, zoom-in and zoom-out, carry different forms of information.}

\subsection{Multi-Stream Feature Integration (MSFI)}
Rather than considering a single stream of information flow, either $S1-S2-S3$ or $S3-S2-S1$, it will be useful to consider multiple streams jointly and to learn features accordingly. With two possible scenarios of `zoom-in' and `zoom-out', we define the information flow of $S1-S2-S3$ as `\emph{zoom-in stream}' and the information flow of $S3-S2-S1$ as `\emph{zoom-out stream}'.  
% Rather than considering a single stream of information flow, either $S1-S2-S3$ or $S3-S2-S1$, it will be useful to consider multiple streams jointly and to learn features accordingly. With two possible scenarios of zoom-in and zoom-out, we define the information flow of S1-S2-S3 as ‘zoom-in stream’ and the information flow of S3-S2-S1 as ‘zoom-out stream’.

As the zoom-in {\HI and} zoom-out streams focus on different scales of morphological and contextual information around the candidate nodule {\HI in a different order}, the learned feature representations from different streams can be complementary to each other {\HI for} FP reduction. Hence, it is desirable to combine such complementary features in a single network {\HI and} to optimize the feature integration from the two streams in an end-to-end manner. To this end, we {\HI design} our network to {\HI integrate contextual information from the two streams as presented in Fig. \ref{fig:MSFR} and call it as} MSFI. 
% As the zoom-in/zoom-out streams focus on different scales of morphological and contextual information around the nodule candidate, the learned feature representations from different streams can be complementary to each other in the task of FP reduction. Hence, it is desirable to combine such complementary features in a single network, thus to optimize the feature representations from the two streams in an end-to-end manner. To this end, we extend our network to accommodate the two streams, and call it as MSFR.
The proposed MSFI is then followed by additional convolutional layers and fully-connected layers to fully define our MGI-CNN, as shown in Fig. \ref{fig:MSFR}.

\begin{table}[tb]
\centering
\renewcommand{\arraystretch}{1.2}
\caption{\HI Statistics of the two datasets, \ie, V1 and V2, for FP reduction in the LUNA16 challenge.  The numbers in parentheses denote the number of nodule-labeled candidates in each dataset that match with the radiologists' decisions.} \label{table:t1}
{\footnotesize
\begin{tabular}{|c|c|c|} \hline
\multirow{2}{*}{Dataset} & \multicolumn{2}{c|}{Candidates} \\\cline{2-3}
            & {Nodule} & {Non-nodule}\\\hline
{V1}        & 1,351 (1,120)    & 549,714 \\ \hline
{V2}        & 1,557 (1,166)    & 753,418 \\ \hline
\end{tabular}
}
\end{table}

\section{Experimental Settings and Results}
\label{sec:experiemtns}

\subsection{Experimental Settings}
\label{subsec:experimental_setting}
We performed the experiments on the LUng Nodule Analysis 2016 (LUNA16) challenge \cite{Setio2017validation} datasets\footnote{\HI Available at `\url{https://luna16.grand-challenge.org}'} by excluding patients whose slice thickness exceeded 2.5 $mm$. LUNA16 includes samples from 888 patients in the LIDC-IDRI open database \cite{Armato2011TheScans.}, which contains annotations of the Ground Truth (GT) collected from the two-step annotation process {\HI by} four experienced radiologists. After each radiologist {\HI annotated} all the candidates on the CT scans, each candidate nodule with the agreement of at least three radiologists {\HI was} approved as GT. There are 1,186 GT nodules in total.
% We performed the experiments on LUng Nodule Analysis 2016 (LUNA16) challenge dataset. The LUNA16 challenge uses samples from 888 patients in the LIDC-IDRI open database \cite{Armato2011TheScans.} by excluding patients, whose slice thickness exceeds 2.5 mm. The LIDC-IDRI database contains annotations of the Ground Truth (GT) collected from the two-step annotation process with four experienced radiologists. 
% After each radiologist annotates all the candidates on  CT scans, the candidate nodule with agree of at least three radiologists approved as GT. There are 1,186 GT nodules in total.
%In the FP reduction track of the LUNA16 challenge, participants submits an algorithm that classifies the false positives of the candidates detected by the candidate nodule detection system into high predictions. 
For the FP reduction challenge, LUNA16 provides {\HI the center coordinates of candidate nodules, the respective patient's ID, and the label information,} obtained by commercially available systems. {\HI Specifically, there are two versions (V1 and V2) of datasets: The V1 dataset provides 551,065 candidate nodules obtained with \cite{Murphy2009AClassification, Tan2011AImages, Jacobs2014AutomaticImages}, of which 1,351 and 549,714 candidates are, respectively, labeled as nodules and non-nodules. In comparison with the four radiologists' decisions, 1,120 nodules out of the 1,351 are matched with GTs}; The V2 dataset includes 754,975 candidate nodules detected with five different nodule detection systems \cite{Murphy2009AClassification, Tan2011AImages, Jacobs2014AutomaticImages, Setio2015AutomaticImages, Traverso2017Computer-aidedChallenges}. {\HI Among the 1,557 nodule-labeled candidates, 1,166 are matched with the GTs, \ie, four radiologists' decisions. 
%In the following, we regard the number of nodule-labeled candidates that
%Of all the candidates, each dataset included only $P$ number of true nodules, while there were $A$ number of true nodules found by radiologists in the original CT images. The numerical values for the candidates are summarized in 
Table \ref{table:t1} summarizes the statistics of the candidate nodules of two datasets for FP reduction in LUNA16.}

\begin{table*}[tb]
\centering 
\caption{\HI Statistics of the training samples used for 5-fold cross-validation.} \label{table:t2}
% {\footnotesize
% \begin{tabular}{\textwidth}{|c|l|c c c c c|} \hline 
\begin{tabular}{|@{\hskip3pt}c@{\hskip3pt}|@{\hskip3pt}l@{\hskip12pt}|c@{\hskip12pt}  c@{\hskip12pt}  c@{\hskip12pt}  c@{\hskip12pt}  c|} \hline 
\multicolumn{2}{|c|}{Dataset}& Fold 1 & Fold 2 & Fold 3 & Fold 4 & Fold 5 \\ \hline \hline 
\multirow{4}{*}{V2} & Scans  & 710 & 710 & 710 & 711 & 711 \\ \cline{2-7} 
{} & Nodule & 1,205 & 1,262 & 1,260 & 1,217 & 1,284 \\ \cline{2-7}
{} & Augmentation & 97,605 & 102,222 & 102,060 & 98,577 & 104,004 \\ \cline{2-7}
{} & Non-nodule & 599,040 & 607,824 & 601,051 & 603,775 & 601,982 \\ \hline 
\multirow{4}{*}{V1} & Scans  & 533 & 532 & 533 & 536 & 530 \\ \cline{2-7}
{} & Nodule & 665 & 677 & 745 & 762 & 709 \\ \cline{2-7}
{} & Augmentation & 53,865 & 54,837 & 60,345 & 61,722 & 57,429 \\ \cline{2-7}
{} & Non-nodule & 330,549 & 330,639 & 330,771 & 330,453 & 330,683 \\ \hline%\hline

\end{tabular}
% }
\end{table*}

\begin{table*}[tb] \centering
\caption{\HI The CPM scores of the competing methods for the FP reduction task on the dataset V2 and V1 in LUNA16 challenge.} 
%(MCNN-RI: Multi-scale CNN (MCNN) with Radical integration of Input patches; MCNN-LR: MCNN with radical integration of Low-level feature Representations; MCNN-ZI: MCNN with zoom-in gradual feature integration; MCNN-ZO: MCNN with zoom-out gradual feature integration, for details refer to the main contexts)}
\label{table:t3}
{\footnotesize
\begin{tabular}{|c|c|c c c c c c c c|} \hline 
\multicolumn{2}{|c|}{} & 0.125 & 0.25 & 0.5 & 1 & 2 & 4 & 8 & Average \\ \hline \hline
\multirow{4}{*}{V2}  
%& MCNN-RI  & 0.887 & 0.921 & 0.939 & 0.943 & 0.947 & 0.958 & 0.962 & 0.936 \\ \cline{2-10} 
%& MCNN-LR  & 0.879 & 0.907 & 0.926 & 0.935 & 0.945 & 0.954 & 0.962 & 0.929 \\ \cline{2-10} 
%& MCNN-ZI  & 0.893 & 0.920 & 0.937 & 0.945 & 0.951 & 0.956 & 0.960 & 0.937 \\ \cline{2-10} 
%& MCNN-ZO  & 0.899 & 0.920 & 0.939 & 0.945 & 0.951 & 0.957 & 0.965 & 0.939 \\ \cline{2-10} 
%& SS-CNN  & 0.902 & 0.919 & 0.934 & 0.941 & 0.950 & 0.960 & 0.964 & 0.938 \\ \cline{2-10} 
%& NS-CNN  & 0.425 & 0.507 & 0.581 & 0.642 & 0.701 & 0.754 & 0.798 & 0.630 \\ \cline{2-10} 
& Proposed MGI-CNN  & \textbf{0.904} & \textbf{0.931} & \textbf{0.943} & 0.947 & 0.952 & 0.956 & 0.962 & \textbf{0.942} \\ \cline{2-10}
& Ding \etal~\cite{Ding2017AccurateNetworks} & 0.797 & 0.857 & 0.895 & 0.938 & 0.954 & 0.970 & 0.981 & 0.913 \\ \cline{2-10}
& Dou \etal~\cite{Dou2016Multi-levelDetection} & 0.677 & 0.834 & 0.927 & \textbf{0.972} & \textbf{0.981} & \textbf{0.983} & \textbf{0.983} & 0.908 \\ \cline{2-10}
& Setio \etal~\cite{Setio2016PulmonaryNetworks} & 0.669 & 0.760  & 0.831 & 0.892 & 0.923 & 0.944 & 0.960 & 0.854 \\ \hline \hline 
% \multirow{4}{*}{V1}& Proposed & \textbf{0.880} & \textbf{0.894} & \textbf{0.907} & \textbf{0.912} & \textbf{0.914} & \textbf{0.919} & \textbf{0.927} & \textbf{0.908} \\ \cline{2-10}
% & Dou \etal~\cite{Dou2016Multi-levelDetection} & 0.678 & 0.738 & 0.816 & 0.848 & 0.879 & 0.907 & 0.922 & 0.827 \\ \cline{2-10}
% & Sakamoto \etal~\cite{Sakamoto2017Multi-stageImages} & 0.760 & 0.794 & 0.833 & 0.860 & 0.876 & 0.893 & 0.906 & 0.846\\ \cline{2-10}
% & Setio \etal~\cite{Setio2016PulmonaryNetworks} & 0.692 & 0.771 & 0.809 & 0.863 & 0.895 & 0.914 & 0.923 & 0.838\\ \hline
% \end{tabular}
% }
% {\footnotesize
% \begin{tabular}{|c|c|c c c c c c c c|} \hline 
% \multicolumn{2}{|c|}{} & 0.125 & 0.25 & 0.5 & 1 & 2 & 4 & 8 & Average \\ \hline \hline
\multirow{4}{*}{V1}& Proposed MGI-CNN & \textbf{0.880} & \textbf{0.894} & \textbf{0.907} & \textbf{0.912} & \textbf{0.914} & \textbf{0.919} & \textbf{0.927} & \textbf{0.908} \\ \cline{2-10}
& Dou \etal~\cite{Dou2016Multi-levelDetection} & 0.678 & 0.738 & 0.816 & 0.848 & 0.879 & 0.907 & 0.922 & 0.827 \\ \cline{2-10}
& Sakamoto \etal~\cite{Sakamoto2017Multi-stageImages} & 0.760 & 0.794 & 0.833 & 0.860 & 0.876 & 0.893 & 0.906 & 0.846\\ \cline{2-10}
& Setio \etal~\cite{Setio2016PulmonaryNetworks} & 0.692 & 0.771 & 0.809 & 0.863 & 0.895 & 0.914 & 0.923 & 0.838\\ \hline
\end{tabular}
}
\end{table*}

% \begin{table*}[tb] \centering
% \renewcommand{\arraystretch}{1.2}
% \caption{The CPM scores of the competing methods for the FP reduction task on the dataset V1 in LUNA16 challenge.}\label{table:t4}
% {\footnotesize
% \begin{tabular}{|c|c|c c c c c c c c|} \hline 
% \multicolumn{2}{|c|}{} & 0.125 & 0.25 & 0.5 & 1 & 2 & 4 & 8 & Average \\ \hline \hline
% \multirow{4}{*}{V1}& Proposed MGI-CNN & \textbf{0.880} & \textbf{0.894} & \textbf{0.907} & \textbf{0.912} & \textbf{0.914} & \textbf{0.919} & \textbf{0.927} & \textbf{0.908} \\ \cline{2-10}
% & Dou \etal~\cite{Dou2016Multi-levelDetection} & 0.678 & 0.738 & 0.816 & 0.848 & 0.879 & 0.907 & 0.922 & 0.827 \\ \cline{2-10}
% & Sakamoto \etal~\cite{Sakamoto2017Multi-stageImages} & 0.760 & 0.794 & 0.833 & 0.860 & 0.876 & 0.893 & 0.906 & 0.846\\ \cline{2-10}
% & Setio \etal~\cite{Setio2016PulmonaryNetworks} & 0.692 & 0.771 & 0.809 & 0.863 & 0.895 & 0.914 & 0.923 & 0.838\\ \hline
% \end{tabular}
% }
% \end{table*}
By using the 3D center coordinates of the candidates provided in the dataset, we extracted a set of 3D patches from thoracic CT scans at scales of $40\times40\times26$, $30\times30\times10$, and $20\times20\times6${\HI, which covered, respectively, 99\%, 85\%, and 58\% of the nodules in the dataset, by following} Dou \etal's work \cite{Dou2016Multi-levelDetection}. The extracted 3D patches were then resized to $20\times20\times6$ {\js by nearest-neighbor interpolation}.
For {\HI faster} convergence, we applied a min-max normalization to patches in the range of [-1000, 400] Hounsfield units (HU)\footnote{a quantitative scale for describing radio density} \cite{Hounsfield1980ComputedImaging}. 

{\HI Regarding the network training, we initialized network parameters with Xavier's method \cite{glorot2010understanding}. We also used a learning rate of 0.003 by decreasing with a weight decay of 2.5\% in every epoch and the number of epochs of 40. For non-linear transformation in convolution and fully-connected layers, we used a ReLU function. To make our network robust, we also applied a dropout technique to fully connected layers with a rate of 0.5. For optimization, we used a stochastic gradient descent with a mini-batch size of 128 and a momentum rate of 0.9.
%{\bck To proposed networks optimization, we use momentum optimizer with 0.9 momentum. } {\color{green}optimizer, regularization} 
}
%We trained our network with the V2 dataset, because more data were available in the V2 dataset. We applied the network trained on the V2 to the V1 dataset and compared the results with other methods.
% Regarding the hyper-paramters for network training, we set a learning rate of 0.003 by decaying to 2.5\%, a batch size of 128, and the number of epochs to 40. We trained our network with a V2 dataset, because the more number of data is available in V2 dataset. For the V1 dataset, we applied our network trained on V2 and reported the results to compare with the competing methods.

% \subsection{Evaluations Metrics}
{\HI For performance evaluation, we used a Competitive Performance Metric (CPM) \cite{Niemeijer2011OnSystems} score, a criterion used in the FP reduction track of the LUNA16 challenge for ranking competitors.
%\textcolor{red}{To false positive reduction track in the LUNA16 challenge, Competitive Performance Metric (CPM) \cite{Niemeijer2011OnSystems} score 
Concretely, a CPM is calculated with 95\% confidence by using bootstrapping \cite{Efron1994AnBootstrap} and averaging sensitivity at seven predefined FP/scan indices, \ie, 0.125, 0.25, 0.5, 1, 2, 4, and 8. For fair comparison with other methods, the performance of our methods reported in this paper were obtained by submitting the probabilities of being nodule for candidate nodules to the website of the LUNA16 challenge. To better justify the validity of the proposed method, we also counted the number nodules and non-nodules correctly classified and thus to present the effect of reducing FPs.}
%Statistical results for the LUNA16 dataset were also verified to confirm the actual number of FP decrements. FP are summarized as one-fold results in each test of 5-fold cross-validation.
% To false positive track in LUNA16 challenge, Competitive Performance Metric (CPM)  score is calculated by averaging sensitivity at seven predefined FP/scan indices: 0.125, 0.25, 0.5, 1, 2, 4, and 8. 
% To samplesets of predefined FP/scan, calculated with 95\% confidence by using bootstrapping \cite{Efron1994AnBootstrap}. Statistical results for the LUNA16 dataset were also verified to confirm the actual number of FP decrements. FP are summarized as one-fold results in each test of 5-fold cross-validation.
% The FP reduction performance is evaluated by detection sensitivity at the average FP/scan. The evaluation result is calculated with 95\% confidence by using bootstrapping \cite{Efron1994AnBootstrap}. A Competitive Performance Metric (CPM) \cite{Niemeijer2011OnSystems} score is calculated by averaging sensitivity at seven predefined FP/scan indices: 0.125, 0.25, 0.5, 1, 2, 4, and 8. 
% The false positive reduction performance is evaluated by the detection sensitivity and the average false positive per scan. The evaluation result is calculated with 95\% confidence by using bootstrapping \cite{Efron1994AnBootstrap}. Competitive Performance Metric (CPM) \cite{Niemeijer2011OnSystems} score is calculated by averaging sensitivity at seven predefined FP/scan: 0.125, 0.25, 0.5, 1, 2, 4, and 8. 

We evaluated the proposed {\HI method} with 5-fold cross-validation. To avoid a potential bias problem {\HI due to the high imbalance in the number of samples between nodules and non-nodules}, we augmented the nodule samples by $ 90^\circ, 180^\circ$, and $270^\circ$ rotation on a transverse plane and 1-pixel shifting along the x, y, and z axes. Consequently, the ratio between the number of nodules to non-nodules was approximately $1:6$. The detailed numbers of nodules and non-nodules are presented in Table \ref{table:t2}.

% The proposed GM-CNN improves performance through the gradual feature extraction strategy and abstract level information utilization strategy with multi-stream feature representation. To empirically identify the effects of each strategy, we tried to verify its performance through different models of similar capacity.
%First, to see the effect of our GFE strategy, we consider two multi-scale CNNs, `Raw-image MCNN (R-MCNN)' and `Low-level MCNN (L-MCNN)', which integrate information from lower layers. R-MCNN uses input as concatenation raw-level patches ($S1 \Vert S2 \Vert S3$) with multi-scale CNN. L-MCNN use the summation of the low-level information extracted from the three multi-scale patches ($F1 \otimes F2 \otimes F3$) as the input to the network.

\subsection{Performance Comparison}% Quantitative analysis of our method
{\HI To verify the validity of the proposed method, \ie, MGI-CNN, we compared with the existing methods \cite{Setio2016PulmonaryNetworks,Dou2016Multi-levelDetection,Ding2017AccurateNetworks,Sakamoto2017Multi-stageImages} in the literature that achieved state-of-the-art performance on V1 and/or V2 datasets of the LUNA16 challenge. Concisely, Setio \etal's method \cite{Setio2016PulmonaryNetworks} uses 9-view 2D patches, Ding \etal's method \cite{Ding2017AccurateNetworks} takes 3D patches as input, and Dou \etal's method \cite{Dou2016Multi-levelDetection} uses multi-level 3D patches. Sakamoto \etal's 2D CNN \cite{Sakamoto2017Multi-stageImages} eliminates the predicted nonconformity in the training data by raising the threshold in every training iteration. 
%Further, to show the effectiveness of our strategies in constructing a multi-scale deep CNN, \ie, GFE in Fig. \ref{fig:GFE} and MSFI in Fig. \ref{fig:MSFR}, we also compared with the following counterpart Multi-scale CNNs (MCCNs): 
%\begin{itemize}
%	\item MCNN with Radical integration of Input patches (MCNN-RI): taking multi-scale 3D patches concatenated at the input-level, \ie, $S1 \Vert S2 \Vert S3$, as input {\color{green}(network architecture?)}
%	\item MCNN with radical integration of Low-level feature Representations (MCNN-LR): integrating multi-scale information at the lower(?) layers {\color{green}(network architecture?)}
%	\item MCNN with zoom-in gradual feature integration (MCNN-ZI): integrating multi-scale patches gradually in the order of $S1-S2-S3$, \ie, the upper network pathway of the proposed network in Fig. \ref{fig:MSFR}
%	\item MCNN with zoom-out gradual feature integration (MCNN-ZO): integrating multi-scale patches gradually in the order of $S3-S2-S1$, \ie, the lower network pathway of the proposed network in Fig. \ref{fig:MSFR}
%%	\item 
%%	\item 
%\end{itemize}
%%Regarding the dataset V2, we compared our performance with state-of-the-art methods in the literature \cite{Setio2016PulmonaryNetworks,Dou2016Multi-levelDetection,Ding2017AccurateNetworks}. Specifically, Setio \etal's method \cite{Setio2016PulmonaryNetworks} uses 9-view 2D patches, Ding \etal's method \cite{Ding2017AccurateNetworks} takes 3D patches as input, and Dou \etal's method \cite{Dou2016Multi-levelDetection} uses multi-level 3D patches. 
Table \ref{table:t3} summarizes the CPM scores over seven different FP/scan values on the V2 and V1 datasets, respectively. 
% Table \ref{table:t3} and Table \ref{table:t4} summarize the CPM scores over seven different FP/scan values on the V2 and V1 datasets, respectively. 
%As for the methods of MCNN-RI, MCNN-LR, MCNN-ZI, and MCNN-ZO, we reported the performance on the V2 dataset only, because the V1 dataset is a subset of the V2 dataset. 

First, on the large-sized V2 dataset, the proposed MGI-CNN was superior to all other competing methods by a large margin in the average CPM.} Notably, when comparing with Dou \etal's method \cite{Dou2016Multi-levelDetection}, which also uses a 3D CNN with the same multi-scale patches as ours, our method increased the average CPM by 0.034 ($\sim$3\% improvement). It is also noteworthy that while the sensitivity of our method at 1, 2, 4, and 8 FP/scan was lower than \cite{Dou2016Multi-levelDetection,Ding2017AccurateNetworks}, our method still achieved the best performance at the 0.125, 0.25, and 0.5 FP/scan. That is, for a low FP rate, which is the main goal of the challenge, our method outperformed those methods.

% In regards to the dataset V2, we compared our performance with the state-of-the-art methods in the literature \cite{Setio2016PulmonaryNetworks,Dou2016Multi-levelDetection,Ding2017AccurateNetworks}. Specifically, Setio \etal's method \cite{Setio2016PulmonaryNetworks} uses 9-view 2D patches, Ding \etal's method \cite{Ding2017AccurateNetworks} takes 3D patches as input, and Dou \etal's method \cite{Dou2016Multi-levelDetection} applies multi-level 3D patches. 
% Table \ref{table:t3} summarizes the CPMs over 7 different FP/scan values over the V2 dataset.
% On average, our GM-CNN was superior to all the competing methods with large margin. 
% Notably, when comparing with Dou \etal's method \cite{Dou2016Multi-levelDetection} that also uses a 3D CNN with the same multi-scale patches as ours, our method increased the average CPM by 0.034 ($\sim$3\% improvement). 
% It is also noteworthy that while the sensitivity of our method at 1, 2, 4, and 8 FP/scan in V2 was lower than \cite{Dou2016Multi-levelDetection,Ding2017AccurateNetworks}, our method still achieved the best performance at the 0.125, 0.25, and 0.5 FP/scan. That is, when considered a low false positive rates, which is the main goal of the challenge, our method outperformed the competing methods. \cite{VanGinneken2015Off-the-shelfScans}

Over the V1 dataset, our method obtained the highest CPMs under all conditions of the FP/scan as presented in Table \ref{table:t3}. Again, when compared with Dou \etal's and Setio \etal~\cite{Setio2016PulmonaryNetworks} \etal's work, our method made promising achievements by increasing the average CPM by 0.081 ($\sim$10\% improvement) and by 0.070 ($\sim$8.35\% improvement). In comparison with Sakamoto \etal's method \cite{Sakamoto2017Multi-stageImages} that reported the highest CPM among the competing methods, our MGI-CNN increased by 0.062 ($\sim$7.3\% improvement). %In this dataset, our method consistently outperformed comparable methods over all FP/scan conditions.

\begin{table*} \centering
\caption{\HI The CPM scores and the number of True Positives (TPs) and False Positives (FPs) of Multi-scale CNNs (MCNN) with different ways of integrating contextual information from input patches. (Radical integration of Input patches; MCNN-LR: MCNN with radical integration of Low-level feature Representations; MCNN-ZI: MCNN with zoom-in gradual feature integration; MCNN-ZO: MCNN with zoom-out gradual feature integration, for details refer to the main contexts)}\label{table:t5}
{\footnotesize
\begin{tabular}{|c|cccccccc|c|c|} \hline 
\multirow{2}{*}{} & \multicolumn{8}{c|}{CPM} & \multirow{2}{*}{TP in GT} & \multirow{2}{*}{FP} \\ \cline{2-9}%& \multirow{2}{*}{FN} \\ \cline{2-9}
 & 0.125 & 0.25 & 0.5 & 1 & 2 & 4 & 8 & Average &    &     \\ \hline \hline%&  \\ \hline \hline
 MCNN-RI  & 0.887 & 0.921 & 0.939 & 0.943 & 0.947 & \textbf{0.958} & 0.962 & 0.936 & 1,159      & 383   \\ \hline%& 119\\ \hline
 MCNN-LR  & 0.879 & 0.907 & 0.926 & 0.935 & 0.945 & 0.954 & 0.962 & 0.929 & 1,156      & 309   \\ \hline%& 109\\ \hline
 MCNN-ZI  & 0.893 & 0.920 & 0.937 & 0.945 & 0.951 & 0.956 & 0.960 & 0.937& 1,160      & 279   \\ \hline%& 106 \\ \hline
 MCNN-ZO  & 0.899 & 0.920 & 0.939 & 0.945 & 0.951 & 0.957 & \textbf{0.965} & 0.939 & \textbf{1,161}      & 267   \\ \hline%& 108\\ \hline
%& NS-CNN  & 0.425 & 0.507 & 0.581 & 0.642 & 0.701 & 0.754 & 0.798 & 0.630 \\ \cline{2-10} 
 Proposed MGI-CNN  & \textbf{0.904} & \textbf{0.931} & \textbf{0.943} & \textbf{0.947} & 0.952 & 0.956 & 0.962 & \textbf{0.942} & \textbf{1,161} & \textbf{232} \\\hline%& \textbf{98} \\\hline
\end{tabular}
}
\end{table*}

%\subsection{Empirical Validation of the Proposed Method} \label{sub:GFE}

\begin{figure*}[t] \centering
%\begin{subfigure}{0.48\textwidth} \centering
\includegraphics[width=1\textwidth, height=2.5cm]{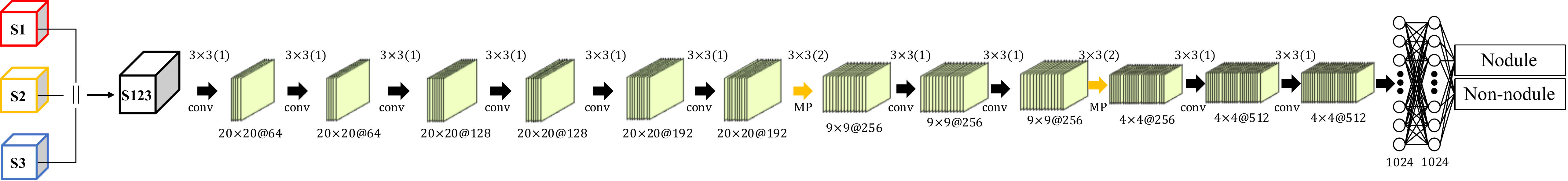}
\caption{\bck Architecture of a multi-scale convolutional neural network with radical integration. {\HI `$\Vert$' denotes concatenation of feature maps. The numbers above the thick black or yellow arrows present a kernel size, \eg, $3\times3$ and a stride, \eg, (1) and (2). (conv: convolution, MP: max-pooling)}}\label{fig:mcnn-ri}
%\end{subfigure}
%\begin{subfigure}{0.48\textwidth} \centering
\includegraphics[width=1\textwidth, height=4cm]{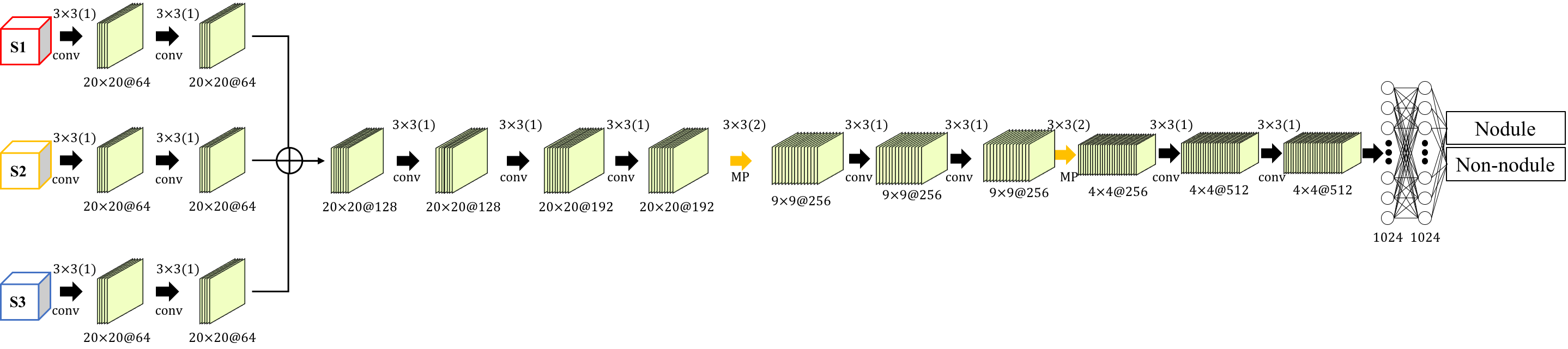}
\caption{\bck Architecture of a multi-scale convolutional neural network with radical integration of low-level feature representations. {\HI `$\oplus$' denotes element-wise summation of feature maps. The numbers above the thick black or yellow arrows present a kernel size, \eg, $3\times3$ and a stride, \eg, (1) and (2). (conv: convolution, MP: max-pooling)}}\label{fig:mcnn-lr}
%\end{subfigure}
%\caption{\bck Examples of the feature maps extracted in the zoom-in/zoom-out stream of the proposed MGI-CNN. The feature maps show gradually extracted contextual information in nodule regions. (a) The zoom-in stream feature maps in the first row show the features of the small-scale patch, and the last row shows the features of the largest scale patch. (b) The zoom-out stream feature maps, on the contrary, show the features of the largest scale patch in the first row and the feature of the smallest scale patch in the last row. \label{fig:mcnn}}
\end{figure*}

{\HI 
\subsection{Effects of the Proposed Strategies}
% \subsubsection{Gradual Feature Extraction} 
To show the effects of our strategies in constructing a multi-scale CNN, \ie, GFE in Fig. \ref{fig:GFE} and MSFI in Fig. \ref{fig:MSFR}, we also conducted experiments with the following Multi-scale CNNs (MCNNs): 
\begin{itemize}
	\item MCNN with Radical integration of Input patches (MCNN-RI): taking multi-scale 3D patches concatenated at the input-level, \ie, $S1 \Vert S2 \Vert S3$, as presented in Fig. \ref{fig:mcnn-ri}. %Table \ref{table:mcnn}.% 
	\item MCNN with radical integration of Low-level feature Representations (MCNN-LR): integrating multi-scale information with feature maps of the first convolution layer as presented in Fig. \ref{fig:mcnn-lr}. %Table \ref{table:mcnn}.%
	\item MCNN with zoom-in gradual feature integration (MCNN-ZI): integrating multi-scale patches gradually in the order of $S1-S2-S3$, \ie, the upper network pathway of the proposed network in Fig. \ref{fig:MSFR}.
	\item MCNN with zoom-out gradual feature integration (MCNN-ZO): integrating multi-scale patches gradually in the order of $S3-S2-S1$, \ie, the lower network pathway of the proposed network in Fig. \ref{fig:MSFR}.
%	\item 
%	\item 
\end{itemize}
To make these networks have similar capacity, we designed network architectures to have a similar number of tunable parameters: MCNN-RI (9,463,320), MCNN-LR (9,466,880), MCNN-ZI (9,464,320), MCNN-ZO (9,464,320), MGI-CNN (9,472,000), where the number of tunable parameters are in parentheses. We conducted this experiment on the V2 dataset only, because the V1 dataset is a subset of the V2 dataset and reported the results in Table \ref{table:t5}. 

First, regarding the strategy of gradual feature extraction, the methods of MCNN-ZI and MCNN-ZO obtained 0.937 and 0.939 of the average CPM, respectively. While the methods with radical integration of contextual information either in the input layer (MCNN-RI) or in the first convolution layer (MCNN-LR) achieved 0.939 and 0.929 of the average CPM. Thus, MCNN-ZI and MCNN-ZO showed slightly higher average CPM scores than MCNN-RI and MCNN-LR. However, in terms of FPs reduction, the power of the gradual feature extraction became notable. That is, while MCNN-RI and MCNN-LR misidentified 383 and 309 non-nodules as nodules, MCNN-ZI and MCNN-ZO failed to remove 279 and 267 non-nodule candidates. 

Second, as for the effect of multi-stream feature integration, the proposed MGI-CNN overwhelmed all the competing methods by achieving the average CPM of 0.942. Further, in FP reduction, MGI-CNN reported only 232 mistakes in filtering out non-nodule candidates. In comparison with MCNN-ZI and MCNN-ZO, the proposed MGI-CNN made 47 and 35 less mistakes, respectively, and thus achieving the best performance in FPs reduction.

It is also worth mentioning that the networks of MCNN-RI, MCNN-LR, MCNN-ZI, MCNN-ZO achieved better performance than the competing methods of \cite{Dou2016Multi-levelDetection,Ding2017AccurateNetworks,Setio2016PulmonaryNetworks} in average CPM. From this comparison, it is believed that the network architectures with the number of tunable parameters of approximately 9.4M had better power of learning feature representations than those of \cite{Dou2016Multi-levelDetection,Ding2017AccurateNetworks,Setio2016PulmonaryNetworks} for FP reduction in pulmonary nodule detection.
}
\begin{table}[t] \centering
\caption{Performance changes of \textcolor{black}{average (Avg.) CPM} according to different stream-integration methods. `TP in GT' denotes the number of true positives that are also included in GT. FP and FN stand for False Positive and False Negative, respectively.} %of \textcolor{red}{Average (Avg.) CPM} 
\label{table:MSFR}
{\footnotesize
\begin{tabular}{|c|c|c|c|c|} \hline
                  & Avg. CPM       & TP in GT    & FP           & FN         \\ \hline
{Concatenation}   & {0.939}        & 1,160       & 263          & 105        \\
{Element-wise sum}& \textbf{0.942} & 1,161       & \textbf{232} & 98  \\
%{Skip connection}& {0.938}        & 1,160       & 335          & 96           \\ 
{$1\times1$ conv} & \textbf{0.942} & 1,160       & 253  & \textbf{93}  \\ \hline
\end{tabular}
}
\end{table}
    
{
Furthermore, the complementary features from the two different streams of GFE should be integrated properly without lowering the performance of FP reduction. To fully utilize the morphological and contextual information while reducing the chance of information loss, we integrate such information with the abstract-level feature representations through MSFI. With an effective integration method, it is possible to compensate for the loss of information that may occur through the feed-forward propagation of the network, especially the max-pooling layer. To combine the feature maps of two streams, we consider three different methods: concatenation, element-wise summation, and $1\times1$ convolution (Table \ref{table:MSFR}). In our experiments, there was no significant difference in the CPM score between element-wise summation and $1\times1$ convolution, but the element-wise summation method achieved the lowest number of FPs, which is the ultimate goal of our work.} %, and element-wise summation with skip connection

\begin{figure*}[t] \centering
\begin{subfigure}{0.48\textwidth} \centering
\includegraphics[width=1\textwidth, height=3cm]{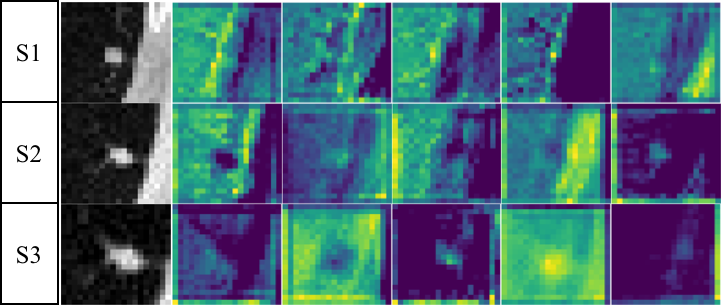}
\caption{\footnotesize {Samples of feature maps from the zoom-in stream}}\label{fig:zIn}
\end{subfigure}
\begin{subfigure}{0.48\textwidth} \centering
\includegraphics[width=1\textwidth, height=3cm]{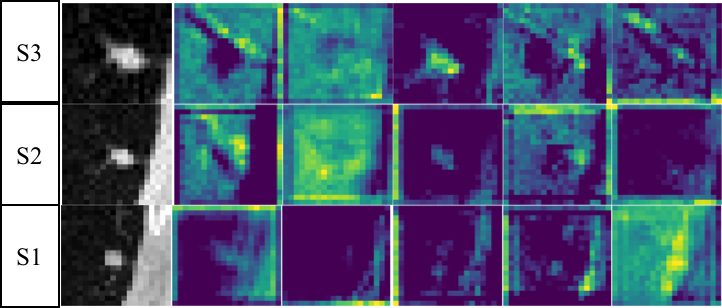}
\caption{\footnotesize {Samples of feature maps from the zoom-out stream}}\label{fig:zOut}
\end{subfigure}
\caption{Examples of the feature maps extracted {\js before concatenation with other scale inputs} in the zoom-in/zoom-out stream of the proposed MGI-CNN. The feature maps show gradually extracted contextual information in nodule regions. (a) The zoom-in stream feature maps in the first row show the features of the small-scale patch, and the last row shows the features of the largest scale patch. (b) The zoom-out stream feature maps, on the contrary, show the features of the largest scale patch in the first row and the feature of the smallest scale patch in the last row. \label{fig:FeatureMap}}
\end{figure*}

\section{Discussions}
\label{sec:discussion}
The major advantages of the proposed method can be summarized by two points. 
First, as shown in Fig. \ref{fig:FeatureMap}, our MGI-CNN could successfully discover morphological and contextual features at different input scales. 
In Fig. \ref{fig:zIn}, we observe that the feature maps in the zoom-in network (\ie, each column in the figure) gradually integrate contextual information in the nodule region. {\js Each sample feature map was extracted from the middle of the sagittal plane in the 3D feature map before concatenation with the next scale input.} A similar but reversed pattern in integrating the contextual information can be observed in the zoom-out network (each column in Fig. \ref{fig:zOut}). These different ways of integrating contextual information and extracting features from multi-scale patches could provide complementary information, and thus could enhance performance in the end. 
Second, our proposed abstract feature integration is useful in terms of information utilization. It is possible to {\js maximize} the FP reduction by integrating features at the abstract-level. 

%skip-connection \cite{xxx}, 
With regard to complementary features integration at the abstract-level, we considered three different strategies, \ie, concatenation, element-wise summation, $1\times1$ convolution \cite{Lin2013network}, commonly used in the literature. The resulting performances are presented in Table \ref{table:MSFR}. Although there is no significant difference among the four methods in average CPM, from a FP reduction perspective, the element-wise summation reported 232 number of FPs, reducing by 31 (vs. concatenation), 103 (vs. skip-connection), and 31 (vs. $1\times1$ convolution). In this regard, we used element-wise summation in our MGI-CNN. 

The 3D patches fed into our network were resized to fit the input receptive field size, \ie, $20\times20\times6$. Such image resizing may cause information loss or corruption in the original-sized patches. However, as we can see in Fig. \ref{fig:FeatureMap}, the 3D patches of size $20\times20\times6$, in which the nodule still occupies most of the patch, was not affected by the resizing operation. This means that even if the surrounding region information is lost by resizing, the information of the nodule itself could be preserved.
% The 3D patches fed into our network were resized to be fit in the input receptive field size, \ie, $[20\times20\times6]$. Such image resizing possibly can  cause information loss or corruption inherent in the original-sized patches. However, as we can see in Fig. \ref{fig:proposed}, the 3D patches with the same size of $[20\times20\times6]$, in which the nodule still occupies most of the patch, was not affected by the resizing operation. This means that even if the surrounding region is lost by resizing, the information of nodule itself could be preserved.

\begin{figure*}[t] \centering
\begin{subfigure}{0.32\textwidth} \centering
\captionsetup{justification=centering}
\includegraphics[width=1\textwidth]{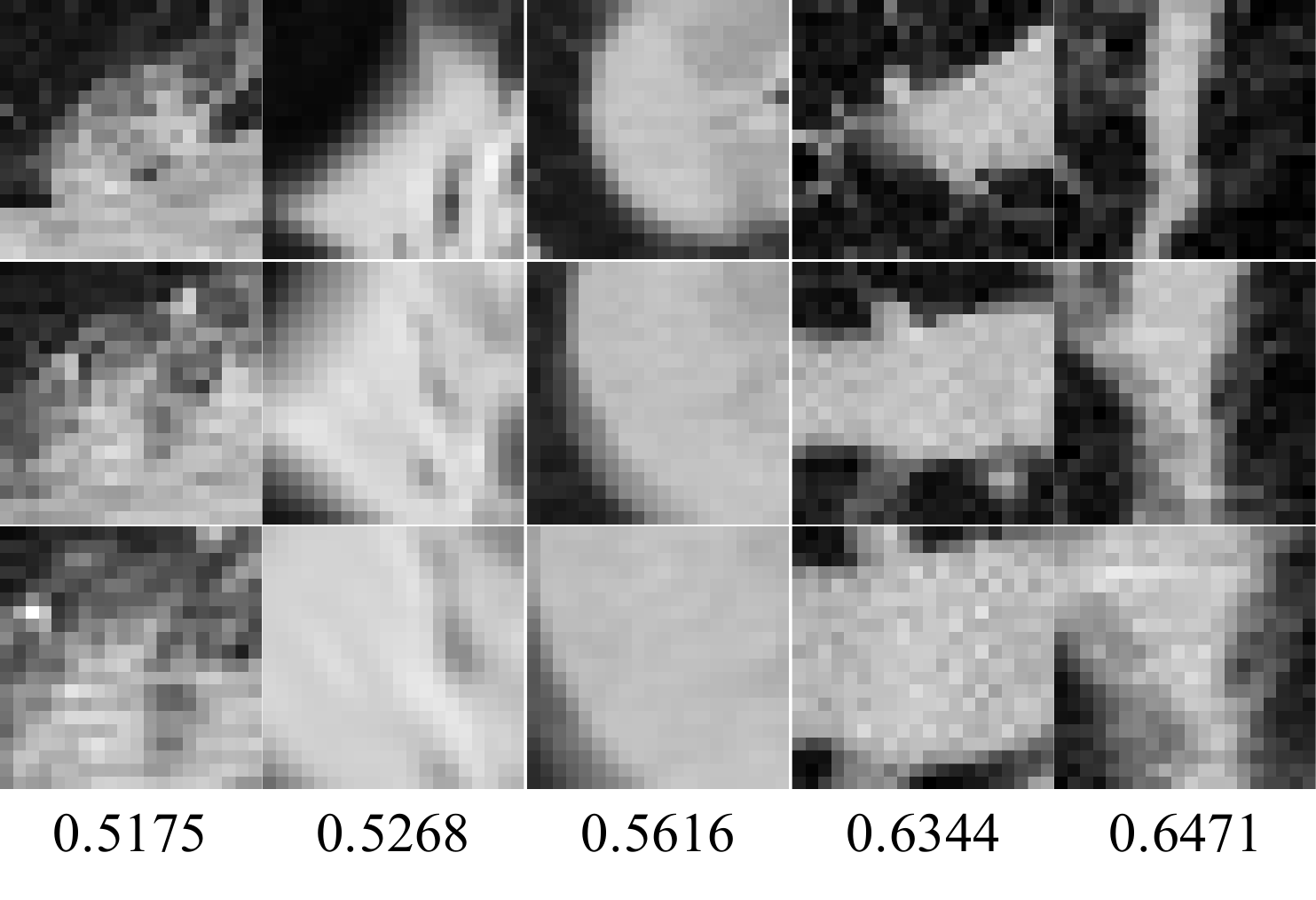}
\caption{ Low confidence \\ ($0.5\leq p <0.7$) }\label{fig:lowp}
\end{subfigure} \centering
\begin{subfigure}{0.32\textwidth} \centering
\captionsetup{justification=centering}
\includegraphics[width=1\textwidth]{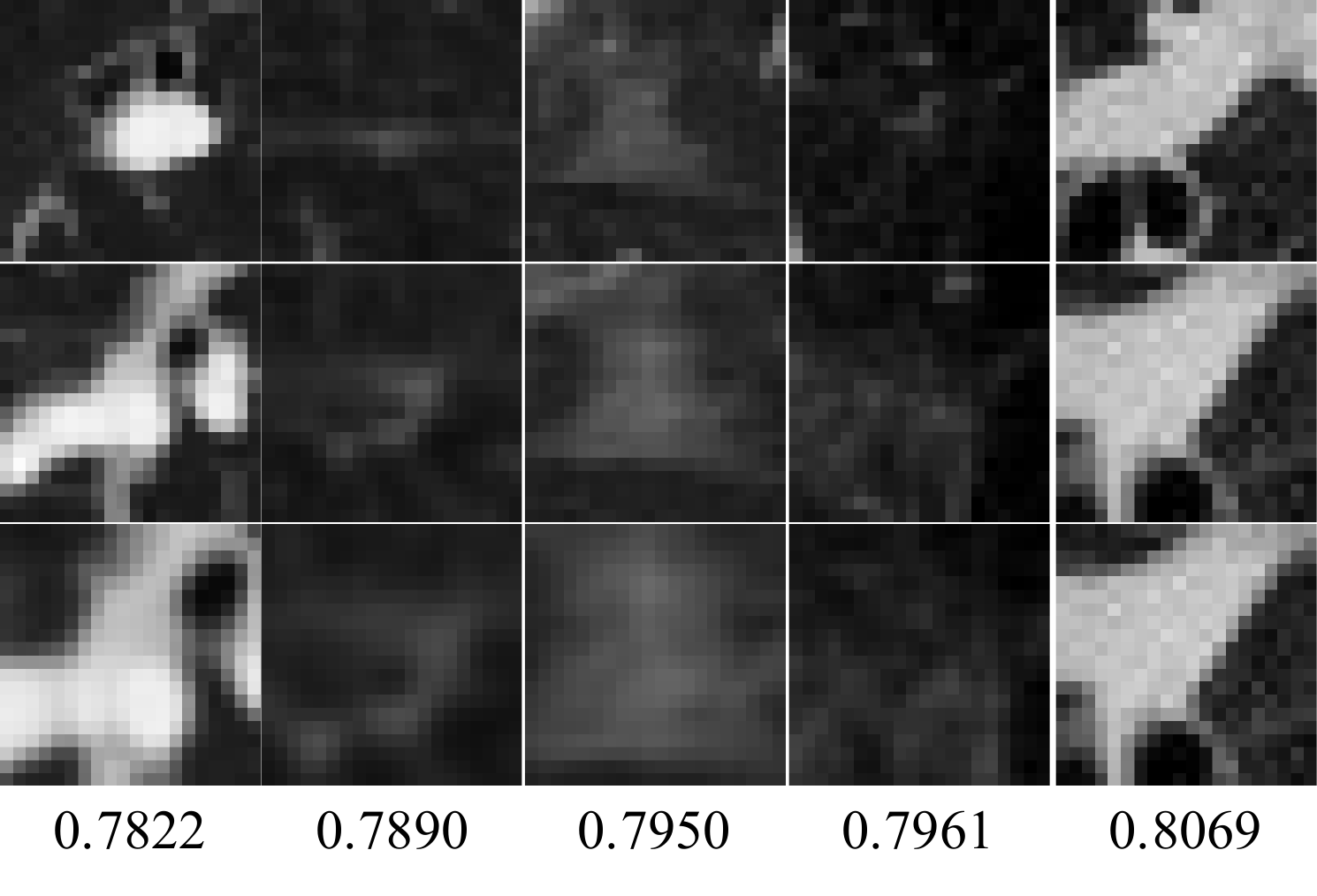} 
\caption{ Moderate confidence \\ ($0.7\leq p <0.9$)}\label{fig:midp}
\end{subfigure}
\begin{subfigure}{0.32\textwidth} \centering
\captionsetup{justification=centering}
\includegraphics[width=1\textwidth]{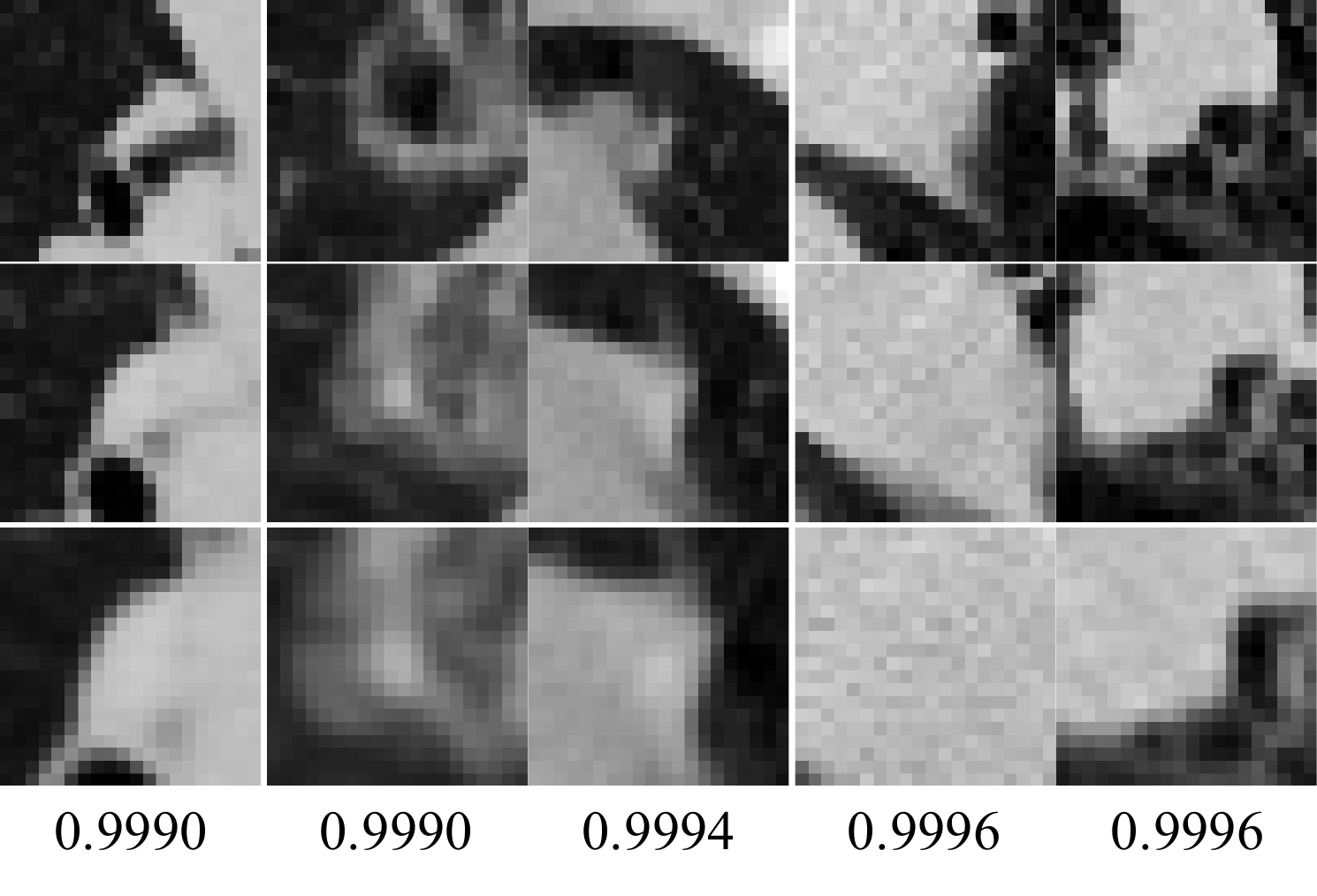}
\caption{ High confidence \\ ($0.9\leq p <1$)}\label{fig:highp}
\end{subfigure}
\caption{Examples of the candidate nodules misclassified to nodule by our MGI-CNN. Based on the output probabilities as nodule, samples are clustered into three groups. The first, second, and third rows correspond to $S1$, $S2$, and $S3$ scale, respectively. The number at the bottom of each column is the output probability as nodule. \label{fig:FPpatch}}

%\includegraphics[width=1\textwidth, height=7cm]{FP_patches.png}
%\caption{ \textcolor{red}{Examples of the FP patch, the information of the candidate is lost and classification is performed based on only the information of the surrounding region and shape features. Given the complex morphological characteristics, errors can occur in the prediction.} \label{fig:FPpatch}}
\end{figure*}

{\HI 
We visually inspected the misclassified candidate nodules. In particular, we first clustered the 232 FPs by our MGI-CNN into three groups based on the their probabilities as nodule: Low Confidence (LC; $0.5\leq p <0.7$), Moderate Confidence (MC; $0.7\leq p<0.9$), and High Confidence (HC; $p>0.9$). The number of FP patches for each group was 33 (LC), 47 (MC), and 152 (HC), respectively. Fig. \ref{fig:FPpatch} presents the representative FP 3D patches for three groups. One noticeable thing from the LC and HC groups is that the extracted 3D patches mostly seem to be a subpart of a large tissue or organ, and thus our network failed to find configural patterns necessary to differentiate from non-nodules. For the MC group, patches show relatively low contrasts, which {\js is} possibly due to our normalization during preprocessing (Section \ref{subsec:experimental_setting}). These observations motivate us to extend our network to accommodate an increased number of patches with larger scales and patches normalized in different ways. This would be an interesting direction to further improve the performance.

From a system's perspective,} instead of developing a full pulmonary nodule detection system, which usually consists of a candidate detection part and a FP reduction part, this study mainly focused on improving the FP reduction component. As the proposed approach is independent of candidate screening methods, our network can be combined with any candidate detector. If the proposed network is combined with more high-performance candidate detection methods, we presume to obtain better results.
% Instead of developing a whole pulmonary nodule detection system, which usually consist of a candidate detection part and a false positive reduction part, this paper mainly focused on improving the false positive reduction component. As the proposed approach is independent of the candidate screening methods, our network can be combined with any candidate detector. If proposed network is combined with more high performance candidate detection methods, we presume to obtain better results.

\section{Conclusion}
\label{sec:conclusion}
In this paper, we proposed a novel {\HI multi-scale gradual integration CNN} for FP reduction in pulmonary nodule detection on thoracic CT scans. {\HI In our network architecture}, we exploited three major strategies: (1) use of multi-scale inputs with different levels of contextual information, (2) gradual integration of the information inherent in different input scales, and (3) multi-stream feature integration by learning in an end-to-end manner. With the first two strategies, we successfully extracted morphological features by gradually integrating contextual information in multi-scale patches. Owing to the third strategy, we could further reduce the number of FPs. In our experiments on the LUNA16 challenge datasets, our network achieved the highest performance with an average CPM of 0.908 on the V1 dataset and an average CPM of 0.942 on the V2 dataset, outperforming state-of-the-art methods by a large margin. In particular, our method obtained promising performances in low FP/scan conditions.
% In this paper, we proposed a novel framework, GM-CNN, for false positive reduction in pulmonary nodule detection on thoracic CT scans. Basically, we exploited three major strategies: (1) use of multi-scale inputs with different levels of contextual information, (2) gradual integration of the information inherent in different input scales, and (3) multi-stream feature representation learning in an end-to-end manner. With the first two strategies, we could successfully extract morphological features by gradually integrating contextual information in multi-scale patches. Due to the third strategy, we could reduce the number of false positives, which empirically validated in our experiments. 
%We qualitativly illustrated the potention of using complementary features discovered from two different network streams, \ie, zoon-in stream and zoom-out stream. We also proposed an effective feature integration method, MSFR, to achieve the best FP reduction performance from the extracted features. 
% In our experiments on the LUNA16 challenge dataset our network achieved the highest average CPMs of 0.908 with the V1 dataset and of 0.942 with the V2 dataset, outperforming the state-of-the-art methods with large margin. In particular, our method obtained promising performances in low FP/scan conditions.

Our current work mostly focused on FP reduction given coordinates of many candidate nodules. We believe that our network can be converted to accomplish positive nodule detection on the low-dose CT scans directly with minor modifications, such as replacing the fully-connected layers with $1\times1$ convolution layers. {\HI For clinical practice, it is also important to classify nodules into various subtypes of solid, non-solid, part-solid, perifissural, calcified, spiculated \cite{Ciompi2017TowardsLearning}, for which different treatments can be used.} 
%{\bck To be used in clinical practice, a pulmonary nodule detection system needs to provide more accurate information. The pulmonary nodules can be subdivided into six types according to their morphological characteristics and their relation to the surroundings: solid, non-solid, part-solid, peri-fissular, calcified, juxta pleural \cite{Ciompi2017TowardsLearning}. Therefore, studies should be conducted to classify various types of nodules rather than simple nodules and non-nodules.}
Thus, it will be our forthcoming research direction.
% Our current work mostly focused on false positive reduction given many nodule candidates. We believe that our network can be converted to accomplish the positive nodule detection directly on the low-dose CT scans with minor modification such as replacing the fully connected layers with 1$\times$1 convolution layers. Thus, it would be our forthcoming research direction.
%Future work
%As the proposed framework has successfully extracted and exploited the contextual features from 3D data, we expect that our network may also be used to be in other object detection tasks in 3D medical image analysis, like Magnetic resonance imaging data in the future work.

% \section*{Acknowledgement}

\section*{Acknowledgment}
This work was supported by Institute for Information \& communications Technology Promotion (IITP) grant funded by the Korea government(MSIT) (No.2017-0-01779, A machine learning and statistical inference framework for explainable artificial intelligence) and also partially supported by Basic Science Research Program through the National Research Foundation of Korea (NRF) funded by the Ministry of Education (NRF-2015R1C1A1A01052216).

{\footnotesize
\bibliographystyle{ieee}
\bibliography{mendeley_ieee}

\begin{thebibliography}{10}\itemsep=-1pt

\bibitem{Armato2011TheScans.}
S.~G. Armato, G.~McLennan, L.~Bidaut, M.~F. McNitt-Gray, C.~R. Meyer, A.~P.
  Reeves, B.~Zhao, D.~R. Aberle, C.~I. Henschke, E.~A. Hoffman, et~al.
\newblock {The Lung Image Database Consortium (LIDC) and Image Database
  Resource Initiative (IDRI): a Completed Reference Database of Lung Nodules on
  CT Scans}.
\newblock {\em Medical Physics}, 38(2):915--931, 2011.

\bibitem{CAO2017327}
P.~Cao, X.~Liu, J.~Yang, D.~Zhao, W.~Li, M.~Huang, and O.~Zaiane.
\newblock A multi-kernel based framework for heterogeneous feature selection
  and over-sampling for computer-aided detection of pulmonary nodules.
\newblock {\em Pattern Recognition}, 64:327--346, 2017.

\bibitem{Ciompi2017TowardsLearning}
F.~Ciompi, K.~Chung, S.~J. van Riel, A.~A.~A. Setio, P.~K. Gerke, C.~Jacobs,
  E.~Th.~Scholten, C.~Schaefer-Prokop, M.~M.~W. Wille, A.~Marchian{\`{o}},
  U.~Pastorino, M.~Prokop, and B.~van Ginneken.
\newblock {Towards Automatic Pulmonary Nodule Management in Lung Cancer
  Screening with Deep Learning}.
\newblock {\em Scientific Reports}, 7(46479), 2017.

\bibitem{Ciompi2015AutomaticOut-of-the-box}
F.~Ciompi, B.~de~Hoop, S.~J. van Riel, K.~Chung, E.~T. Scholten, M.~Oudkerk,
  P.~A. de~Jong, M.~Prokop, and B.~van Ginneken.
\newblock {Automatic Classification of Pulmonary Peri-fissural Nodules in
  Computed Tomography using an Ensemble of 2D Views and a Convolutional Neural
  Network Out-of-The-Box}.
\newblock {\em Medical Image Analysis}, 26(1):195--202, 2015.

\bibitem{Ding2017AccurateNetworks}
J.~Ding, A.~Li, Z.~Hu, and L.~Wang.
\newblock {Accurate Pulmonary Nodule Detection in Computed Tomography Images
  Using Deep Convolutional Neural Networks}.
\newblock {\em arXiv preprint arXiv:1706.04303}, pages 1--9, 2017.

\bibitem{Dou2016Multi-levelDetection}
Q.~Dou, H.~Chen, L.~Yu, J.~Qin, and P.~A. Heng.
\newblock {Multi-level Contextual 3D CNNs for False Positive Reduction in
  Pulmonary Nodule Detection}.
\newblock {\em IEEE Transactions on Biomedical Engineering}, 64(7):1558--1567,
  2017.

\bibitem{Dou2016AutomaticNetworks}
Q.~Dou, H.~Chen, L.~Yu, L.~Zhao, J.~Qin, D.~Wang, V.~C. Mok, L.~Shi, and P.~A.
  Heng.
\newblock {Automatic Detection of Cerebral Microbleeds From MR Images via 3D
  Convolutional Neural Networks}.
\newblock {\em IEEE Transactions on Medical Imaging}, 35(5):1182--1195, 2016.

\bibitem{Efron1994AnBootstrap}
B.~Efron and R.~Tibshirani.
\newblock {\em {An Introduction to the Bootstrap}}.
\newblock Chapman {\&} Hall, 1994.

\bibitem{Esteva2017Dermatologist-levelNetworks}
A.~Esteva, B.~Kuprel, R.~A. Novoa, J.~Ko, S.~M. Swetter, H.~M. Blau, and
  S.~Thrun.
\newblock {Dermatologist-Level Classification of Skin Cancer with Deep Neural
  Networks}.
\newblock {\em Nature}, 542(7639):115--118, 2017.

\bibitem{Friedman1998AdditiveBoosting}
J.~Friedman, T.~Hastie, R.~Tibshirani, and Y.~Stanford.
\newblock {Additive Logistic Regression: a Statistical View of Boosting}.
\newblock {\em The Annals of Statistics}, 28(2):337--407, 1998.

\bibitem{glorot2010understanding}
X.~Glorot and Y.~Bengio.
\newblock {Understanding The Difficulty of Training Deep Feedforward Neural
  Networks}.
\newblock In {\em In Proceedings of the International Conference on Artificial
  Intelligence and Statistics}, pages 249--256, 2010.

\bibitem{Gould2007EvaluationEdition}
M.~K. Gould, J.~Fletcher, M.~D. Iannettoni, W.~R. Lynch, D.~E. Midthun, D.~P.
  Naidich, and D.~E. Ost.
\newblock {Evaluation of Patients with Pulmonary Nodules: When is it Lung
  Cancer? ACCP Evidence-Based Clinical Practice Guidelines (2nd edition)}.
\newblock {\em Chest}, 132(3):108S--130S, 2007.

\bibitem{Havaei2017BrainNetworks}
M.~Havaei, A.~Davy, D.~Warde-Farley, A.~Biard, A.~Courville, Y.~Bengio, C.~Pal,
  P.~M. Jodoin, and H.~Larochelle.
\newblock {Brain Tumor Segmentation with Deep Neural Networks}.
\newblock {\em Medical Image Analysis}, 35:18--31, 2017.

\bibitem{Honari2016RecombinatorAggregation}
S.~Honari, J.~Yosinski, P.~Vincent, and C.~Pal.
\newblock {Recombinator Networks: Learning Coarse-to-Fine Feature Aggregation}.
\newblock In {\em Proceedings of the IEEE Conference on Computer Vision and
  Pattern Recognition}, pages 1--11, 2016.

\bibitem{Hounsfield1980ComputedImaging}
G.~Hounsfield.
\newblock {Computed Medical Imaging}.
\newblock {\em Science}, 210(4465), 1980.

\bibitem{HU2018134}
Z.~Hu, J.~Tang, Z.~Wang, K.~Zhang, L.~Zhang, and Q.~Sun.
\newblock Deep learning for image-based cancer detection and diagnosis - a
  survey.
\newblock {\em Pattern Recognition}, 83:134--149, 2018.

\bibitem{Jacobs2014AutomaticImages}
C.~Jacobs, E.~M. van Rikxoort, T.~Twellmann, E.~T. Scholten, P.~A. de~Jong,
  J.-M. Kuhnigk, M.~Oudkerk, H.~J. de~Koning, M.~Prokop, C.~Schaefer-Prokop,
  et~al.
\newblock {Automatic Detection of Subsolid Pulmonary Nodules in Thoracic
  Computed Tomography Images}.
\newblock {\em Medical Image Analysis}, 18(2):374--384, 2014.

\bibitem{Kamnitsas2017EfficientSegmentation}
K.~Kamnitsas, C.~Ledig, V.~F. Newcombe, J.~P. Simpson, A.~D. Kane, D.~K. Menon,
  D.~Rueckert, and B.~Glocker.
\newblock {Efficient multi-scale 3D CNN with fully connected CRF for accurate
  brain lesion segmentation}.
\newblock {\em Medical Image Analysis}, 36:61--78, 2017.

\bibitem{Karpathy2014Large-ScaleNetworks}
A.~Karpathy, G.~Toderici, S.~Shetty, T.~Leung, R.~Sukthankar, and L.~Fei-Fei.
\newblock {Large-Scale Video Classification with Convolutional Neural
  Networks}.
\newblock In {\em Proceedings of the IEEE Conference on Computer Vision and
  Pattern Recognition}, pages 1725--1732, 2014.

\bibitem{Lee2001AutomatedTechnique}
Y.~Lee, T.~Hara, H.~Fujita, S.~Itoh, and T.~Ishigaki.
\newblock {Automated Detection of Pulmonary Nodules in Helical CT Images Based
  on an Improved Template-Matching Technique}.
\newblock {\em IEEE Transactions on Medical Imaging}, 20(7):595--604, 2001.

\bibitem{Li2003SelectiveScans}
Q.~Li, S.~Sone, and K.~Doi.
\newblock {Selective Enhancement Filters for Nodules, Vessels, and Airway Walls
  in Two- and Three-Dimensional CT Scans}.
\newblock {\em Medical Physics}, 30(8):2040--2051, 2003.

\bibitem{Lin2013network}
M.~Lin, Q.~Chen, and S.~Yan.
\newblock {Network in Network}.
\newblock {\em arXiv preprint arXiv:1312.4400}, 2013.

\bibitem{Lin2016FeatureDetection}
T.-Y. Lin, P.~Doll{\'a}r, R.~Girshick, K.~He, B.~Hariharan, and S.~Belongie.
\newblock {Feature Pyramid Networks for Object Detection}.
\newblock {\em arXiv preprint arXiv:1612.03144}, 2016.

\bibitem{LIU2018262}
X.~Liu, F.~Hou, H.~Qin, and A.~Hao.
\newblock Multi-view multi-scale cnns for lung nodule type classification from
  ct images.
\newblock {\em Pattern Recognition}, 77:262--275, 2018.

\bibitem{Murphy2009AClassification}
K.~Murphy, B.~van Ginneken, A.~M. Schilham, B.~De~Hoop, H.~Gietema, and
  M.~Prokop.
\newblock {A Large-Scale Evaluation of Automatic Pulmonary Nodule Detection in
  Chest CT using Local Image Fatures and k-Nearest-Neighbour Classification}.
\newblock {\em Medical Image Analysis}, 13(5):757--770, 2009.

\bibitem{Nair2010RectifiedMachines}
V.~Nair and G.~E. Hinton.
\newblock {Rectified Linear Units Improve Restricted Boltzmann Machines}.
\newblock In {\em Proceedings of International Conference on Machine Learning},
  pages 807--814, 2010.

\bibitem{NationalLungScreeningTrialResearchTeam2011ReducedScreening}
{National Lung Screening Trial Research Team}, D.~R. Aberle, A.~M. Adams, C.~D.
  Berg, W.~C. Black, J.~D. Clapp, R.~M. Fagerstrom, I.~F. Gareen, C.~Gatsonis,
  P.~M. Marcus, and J.~D. Sicks.
\newblock {Reduced Lung-Cancer Mortality with Low-Dose Computed Tomographic
  Screening}.
\newblock {\em New England Journal of Medicine}, 365(5):395--409, 8 2011.

\bibitem{Niemeijer2011OnSystems}
M.~Niemeijer, M.~Loog, M.~D. Abramoff, M.~A. Viergever, M.~Prokop, and B.~van
  Ginneken.
\newblock {On Combining Computer-Aided Detection Systems}.
\newblock {\em IEEE Transactions on Medical Imaging}, 30(2):215--223, 2011.

\bibitem{Okumura1998AutomaticFilter}
T.~Okumura, T.~Miwa, J.-I. Kako, S.~Yamamoto, R.~Matsumoto, Y.~Tateno,
  T.~Iinuma, and T.~Matsumoto.
\newblock {Automatic Detection of Lung Cancers in Chest CT Images by Variable
  N-Quoit Filter}.
\newblock In {\em Proceedings of International Conference on Pattern
  Recognition}, volume~2, pages 1671--1673, 1998.

\bibitem{Razavian2014}
A.~S. Razavian, H.~Azizpour, J.~Sullivan, and S.~Carlsson.
\newblock {CNN Features Off-the-Shelf: An Astounding Baseline for Recognition}.
\newblock In {\em Proceedings of the IEEE Conference on Computer Vision and
  Pattern Recognition Workshops}, pages 512--519, 2014.

\bibitem{Roth2016ImprovingAggregation}
H.~R. Roth, L.~Lu, J.~Liu, J.~Yao, A.~Seff, K.~Cherry, L.~Kim, and R.~M.
  Summers.
\newblock {Improving Computer-Aided Detection using Convolutional Neural
  Networks and Random View Aggregation}.
\newblock {\em IEEE Transactions on Medical Imaging}, 35(5):1170--1181, 2016.

\bibitem{Sakamoto2017Multi-stageImages}
M.~Sakamoto, H.~Nakano, K.~Zhao, and T.~Sekiyama.
\newblock {Multi-stage Neural Networks with Single-Sided Classifiers for False
  Positive Reduction and Its Evaluation Using Lung X-Ray (CT) Images}.
\newblock In {\em Proceedings of International Conference Image Analysis and
  Processing}, pages 370--379, 2017.

\bibitem{Setio2016PulmonaryNetworks}
A.~A.~A. Setio, F.~Ciompi, G.~Litjens, P.~Gerke, C.~Jacobs, S.~J. Van~Riel,
  M.~M. Winkler~Wille, M.~Naqibullah, C.~I. S{\'{a}}nchez, and B.~Van~Ginneken.
\newblock {Pulmonary Nodule Detection in CT Images: False Positive Reduction
  Using Multi-View Convolutional Networks}.
\newblock {\em IEEE Transactions on Medical Imaging}, 35(5):1160--1169, 2016.

\bibitem{Setio2015AutomaticImages}
A.~A.~A. Setio, C.~Jacobs, J.~Gelderblom, and B.~van Ginneken.
\newblock {Automatic Detection of Large Pulmonary Solid Nodules in Thoracic CT
  Images}.
\newblock {\em Medical Physics}, 42(10):5642--5653, 2015.

\bibitem{Setio2017validation}
A.~A.~A. Setio, A.~Traverso, T.~De~Bel, M.~S. Berens, C.~van~den Bogaard,
  P.~Cerello, H.~Chen, Q.~Dou, M.~E. Fantacci, B.~Geurts, et~al.
\newblock {Validation, Comparison, and Combination of Algorithms for Automatic
  Detection of Pulmonary Nodules in Computed Tomography Images: The LUNA16
  Challenge}.
\newblock {\em Medical Image Analysis}, 42:1--13, 2017.

\bibitem{Shen2017DeepAnalysis.}
D.~Shen, G.~Wu, and H.-I. Suk.
\newblock {Deep Learning in Medical Image Analysis.}
\newblock {\em Annual Review of Biomedical Engineering}, 19:221--248, 2017.

\bibitem{Shen2015}
W.~Shen, M.~Zhou, F.~Yang, C.~Yang, and J.~Tian.
\newblock {Multi-scale Convolutional Neural Networks for Lung Nodule
  Classification}.
\newblock In {\em Proceedings of International Conference on Information
  Processing in Medical Imaging}, pages 588--599. Springer, 2015.

\bibitem{SHEN2017663}
W.~Shen, M.~Zhou, F.~Yang, D.~Yu, D.~Dong, C.~Yang, Y.~Zang, and J.~Tian.
\newblock Multi-crop convolutional neural networks for lung nodule malignancy
  suspiciousness classification.
\newblock {\em Pattern Recognition}, 61:663--673, 2017.

\bibitem{Shen2017Multi-cropClassification}
W.~Shen, M.~Zhou, F.~Yang, D.~Yu, D.~Dong, C.~Yang, Y.~Zang, and J.~Tian.
\newblock {Multi-crop Convolutional Neural Networks for Lung Nodule Malignancy
  Suspiciousness Classification}.
\newblock {\em Pattern Recognition}, 61:663--673, 2017.

\bibitem{Shin2016DeepLearning}
H.~C. Shin, H.~R. Roth, M.~Gao, L.~Lu, Z.~Xu, I.~Nogues, J.~Yao, D.~Mollura,
  and R.~M. Summers.
\newblock {Deep Convolutional Neural Networks for Computer-Aided Detection: CNN
  Architectures, Dataset Characteristics and Transfer Learning}.
\newblock {\em IEEE Transactions on Medical Imaging}, 35(5):1285--1298, 2016.

\bibitem{Siegel2017Cancer2017}
R.~L. Siegel, K.~D. Miller, and A.~Jemal.
\newblock {Cancer Statistics, 2017}.
\newblock {\em CA: A Cancer Journal for Clinicians}, 67(1):7--30, 1 2017.

\bibitem{Suzuki2003MassiveTomography}
K.~Suzuki, S.~G. Armato, F.~Li, S.~Sone, and K.~Doi.
\newblock {Massive Training Artificial Neural Network (MTANN) for Reduction of
  False Positives in Computerized Detection of Lung Nodules in Low-Dose
  Computed Tomography}.
\newblock {\em Medical Physics}, 30(7):1602--1617, 2003.

\bibitem{Tan2011AImages}
M.~Tan, R.~Deklerck, B.~Jansen, M.~Bister, and J.~Cornelis.
\newblock {A Novel Computer-Aided Lung Nodule Detection System for CT Images}.
\newblock {\em Medical Physics}, 38(10):5630--5645, 2011.

\bibitem{Traverso2017Computer-aidedChallenges}
A.~Traverso, E.~L. Torres, M.~E. Fantacci, and P.~Cerello.
\newblock {Computer-Aided Detection Systems to Improve Lung Cancer Early
  Diagnosis: State-of-the-art and Challenges}.
\newblock In {\em Proceedings of Journal of Physics: Conference Series}, volume
  841, page 012013, 2017.

\bibitem{Yang2009_TL}
Q.~Yang and S.~J. Pan.
\newblock {A Survey on Transfer Learning}.
\newblock {\em IEEE Transactions on Knowledge and Data Engineering},
  22:1345--1359, 2009.

\bibitem{Ye2009Shape-basedImages}
X.~Ye, X.~Lin, J.~Dehmeshki, G.~Slabaugh, and G.~Beddoe.
\newblock {Shape-Based Computer-Aided Detection of Lung Nodules in Thoracic CT
  Images}.
\newblock {\em IEEE Transactions on Biomedical Engineering}, 56(7):1810--1820,
  2009.

\bibitem{Zhang2014ScaleAnalysis}
J.~Zhang, P.~M. Atkinson, and M.~F. Goodchild.
\newblock {\em {Scale in Spatial Information and Analysis}}.
\newblock CRC Press, Taylor and Francis, 2014.

\end{thebibliography}
}

\end{document}